\documentclass[lettersize,journal]{IEEEtran}
\usepackage{amsmath,amsfonts}
\usepackage{ amssymb }
\usepackage[ruled]{algorithm2e}
\usepackage{array}
\usepackage[caption=false,font=normalsize,labelfont=sf,textfont=sf]{subfig}
\usepackage{textcomp}
\usepackage{stfloats}
\usepackage{url}

\usepackage{verbatim}
\usepackage{graphicx}
\usepackage{cite}
\usepackage{multirow}
\usepackage{threeparttable}
\usepackage{nomencl}
\usepackage{color}
\usepackage{booktabs}
\makenomenclature
\newtheorem{Definition}{Definition}
\usepackage{bm}

\DeclareMathAlphabet{\mathbb}{U}{bbold}{m}{n}

\begin{document}

\title{Neural Influence Estimator: Towards Real-time Solutions to Influence Blocking Maximization}
\author{Wenjie Chen,
		Shengcai Liu*,~\IEEEmembership{Member,~IEEE},
		Yew-Soon Ong,~\IEEEmembership{Fellow,~IEEE},
            Li Zhuang,
		Ke Tang,~\IEEEmembership{Fellow,~IEEE}
	
	\thanks{
		Manuscript received (date to be filled in by Editor). This work was supported by Guangdong Major Project of Basic and Applied Basic Research under Grant 2023B0303000010, National Natural Science Foundation of China under Grant 72401105, Hubei Provincial Natural Science Foundation of China under Grant 2024AFB338, “Zhishan” Scholars Programs of Southeast University under Grant 2242023R40056, the Distributed Smart Value Chain programme which is funded under the Singapore RIE2025 Manufacturing, Trade and Connectivity (MTC) Industry Alignment Fund-Pre-Positioning (Award No: M23L4a0001), the MTI under its AI Centre of Excellence for Manufacturing (AIMfg) (Award No: W25MCMF014), and the College of Computing and Data Science, Nanyang Technological University.
		
		Wenjie Chen is with the School of Information Management, Central China Normal University, Wuhan, China. 
		
		Shengcai Liu and Ke Tang are with the Guangdong Provincial Key Laboratory of Brain-Inspired Intelligent Computation, Department of Computer Science and Engineering, Southern University of Science and Technology, Shenzhen, China. 
		
		Yew-Soon Ong is with the Centre for Frontier AI Research, Institute of High Performance Computing, Agency for Science, Technology and Research, Singapore, and the College of Computing and Data Science, Nanyang Technological University.

        Li Zhuang is with the School of Cyber Science and Engineering, Southeast University, Nanjing, China.  
		
		*Corresponding Author. Email: liusc3@sustech.edu.cn
}}

\markboth{IEEE TRANSACTIONS ON }%
{Shell \MakeLowercase{\textit{CHEN et al.}}: Neural Influence Estimator: Towards Real-time Solutions to Influence Blocking Maximization}

\maketitle

\begin{abstract}
Real-time solutions to the influence blocking maximization (IBM) problems are crucial for promptly containing the spread of misinformation.
However, achieving this goal is non-trivial, mainly because assessing the blocked influence of an IBM problem solution typically requires plenty of expensive Monte Carlo simulations (MCSs).
This work presents a novel approach that enables solving IBM problems with hundreds of thousands of nodes and edges in seconds.
The key idea is to construct a fast-to-evaluate surrogate model called neural influence estimator (NIE) offline as a substitute for the time-intensive MCSs, and then combine it with optimization algorithms to address IBM problems online.
To this end, a learning problem is formulated to build the NIE that takes the false-and-true information instance as input, extracts features describing the topology and inter-relationship between two seed sets, and predicts the blocked influence.
A well-trained NIE can generalize across different IBM problems given a social network, and can be readily combined with existing IBM optimization algorithms.
The experiments on 25 IBM problems with up to millions of edges show that the NIE-based optimization method can be up to four orders of magnitude faster than MCSs-based optimization method to achieve the same optimization quality.
Moreover, given a one-minute limit, the NIE-based method can solve IBM problems with up to hundreds of thousands of nodes, which is at least one order of magnitude larger than what can be solved by existing methods.
\end{abstract}


\section{Introduction}
\label{Introduction}

\IEEEPARstart{O}{nline} social networks have become vital communication platforms for users; however, they have also emerged as breeding grounds for the spread of rumors and hate speech.
According to World Economic Forum~\cite{WEF}, misinformation on social media has been recognized as one of the global risks. 
One common approach to mitigating the impact of misinformation is to distribute correct information to raise users' awareness and thus limit the spread of misinformation.
This approach is commonly known as influence blocking maximization (IBM)~\cite{chen2022influence,yang2024deadline,ni2024minimizing,gao2024efficient}.

In practice, misinformation, when spread explosively, has the potential to severely disrupt public order and social stability.
An illustrative case~\cite{HS} took place in 2017 when the U.S. National Weather Service issued inaccurate information about the Oroville Dam evacuation.
Numerous emergency calls quickly flooded the 911 dispatch center and occupied a significant portion of emergency resources for an extended time period.
As a consequence of the delayed response, genuine emergency calls were left unattended during that time.
Another example~\cite{UN} is that a community in Uganda propagated rumors that attributed the cause of COVID-19 to evil spirits.
Consequently, hate speech against foreigners surged, triggering serious social unrest in Uganda.

To prevent such vicious incidents, it is crucial to promptly address and contain misinformation before it spreads widely.
From the problem-solving perspective, this necessitates a method capable of real-time solutions (i.e., responses within a minute) to practical IBM problems with hundreds of thousands of nodes and beyond.
However, attaining this goal is non-trivial.
Firstly, the IBM problem is an NP-hard combinatorial optimization problem~\cite{budak2011limiting}, meaning that one would typically resort to approximation and heuristic algorithms, such as the greedy algorithm, to achieve low time complexity.
Secondly and more importantly, assessing the solution quality (i.e., the blocked influence) is a \#P-hard problem~\cite{chen2010scalable1}, which requires plenty of expensive Monte Carlo simulations (MCSs).

Unfortunately, despite the pressing needs, existing methods for solving IBM problems often fail to provide timely solutions.
Budak et al.~\cite{budak2011limiting} conducted pioneering research on solving IBM problems and proposed an MCSs-based greedy algorithm.
However, this algorithm would take approximately 16 hours to solve the IBM problems with around 6,000 nodes.
Subsequently, efforts have been made to reduce the number of MCSs required to solve IBM problems or to reduce the computational cost of each simulation.
For example, He et al.~\cite{he2022reinforcement} proposed to use a candidate set to reduce the search space, which identified a good solution with fewer MCSs compared with searching the entire solution space.
Yang and Li~\cite{yang2024deadline} introduced a sketch-based optimization method that avoids repetitive MCSs.
Overall, these methods still demand a significant amount of time to solve IBM problems with thousands of nodes.

This work aims to develop a novel approach that can achieve real-time solutions to IBM problems of practical scales.
Specifically, our interest lies in addressing IBM problems that encompass hundreds of thousands of nodes and edges, with the goal of finding good solutions within seconds (less than one minute).
The key idea is to construct a fast-to-evaluate surrogate model offline as a substitute for the time-intensive MCSs.
To this end, a learning problem is formulated to build the neural influence estimator (NIE).
The NIE takes the false-and-true information instance, which contains a false-information
seed set and a true-information seed set, as input.
Then, it extracts features describing the topology and inter-relationship between these two seed sets and predicts the blocked inﬂuence.

Such a design enables a well-trained NIE to generalize across different IBM problems with varying false-and-true information instances on  a given social network.
The time complexity of NIE is remarkably low, making it highly scalable and capable of handling large-scale IBM problems effortlessly.
As a solution evaluation approach, NIE can be readily combined with existing IBM optimization algorithms such as the greedy algorithms and global optimizers.

The main contributions are summarized as follows:
\begin{enumerate}
	\item A novel method is proposed to facilitate real-time solutions for IBM problems at practical scales. It introduces a fast influence estimator NIE to replace the expensive simulation evaluations, and integrates NIE with optimization algorithms to enhance efficiency.
	\item A learning problem is formulated to develop NIE, which is capable of predicting the blocked influence of any false-and-true information instance given a social network. This involves a novel feature extraction method that considers both the network's topology and the relationships between false-information and true-information seed sets. Complexity analysis indicates that NIE has significantly lower time complexity compared to existing influence estimators, enabling it to efficiently scale to larger networks.
	\item Experimental results on 25 IBM problems with up to millions of edges verify that the NIE-based optimization method significantly outperforms the state-of-the-art methods in terms of both runtime and quality, given a one-minute time limit.
	Notably, the NIE-based optimization method can be up to four orders of magnitude faster than the MCSs-based optimization methods to achieve the same solution quality.
	Moreover, given the constraint of one minute, the NIE-based method can solve IBM problems with hundreds of thousands of nodes, which is at least one order of magnitude larger than what can be solved by existing methods.
\end{enumerate}

The remainder of the paper is organized as follows.
Section~\ref{Related Works} reviews the related works on improving the efficiency of the MCSs-based greedy algorithm for the IBM problems.
In Section~\ref{Problem Defination and Notations}, the information diffusion model and the IBM problem are introduced.
Section~\ref{Neural Network-based Surrogate Model} presents NIE, analyzes time complexity, and introduces an NIE-based optimization method.
Experimental results are presented in Section~\ref{Experimental Studies}, followed by the conclusion in Section~\ref{Conclusion}.
	
\section{Related Works}
\label{Related Works}

As aforementioned, the MCSs-based greedy algorithm is inefficient in tackling IBM problems. 
This section reviews related works on improving the efficiency of the MCSs-based greedy algorithm for the IBM problems from two aspects: reducing the number of evaluations and cutting the expense of a single evaluation.

\subsection{Reducing the Number of Evaluations}
Both the speed and the number of evaluations can affect optimization efficiency.
Some studies attempt to achieve a good solution with fewer influence estimations.
Zhang et al.~\cite{zhang2015limiting} identified the gateway nodes that enlarged the spread of misinformation to construct the candidate set and used the greedy algorithm to identify the final solution.
Yan et al.~\cite{yan2019minimizing} proposed a two-stage method that initially chose the candidate nodes with strong ability to block misinformation and then applied the greedy algorithm to the candidate set.
He et al.~\cite{he2022reinforcement} partitioned the network into multiple communities, selected candidate nodes from each community, and used reinforcement learning to identify high-quality solutions.
Although reducing the number of evaluations is not the focus of this work, these techniques can be combined with our approach with ease to further enhance efficiency.

\subsection{Accelerating a Single Evaluation}
The first category of acceleration methods involves limiting the scope or path of influence spread.
He et al.~\cite{he2012influence} picked node's local subgraph based on the directed acyclic graphs and employed a dynamic programming method to compute its influence within the subgraph. 
Wu and Pan~\cite{wu2017scalable} utilized the maximum influence arborescence structure to restrict the propagation of influence along the path with the highest propagation probability, thus reducing the simulation's time complexity.
However, the limitation of these methods is their susceptibility to the graph structure, which affects the effectiveness of acceleration.

The second way of reducing evaluation expenses is to avoid rerunning simulations, known as the sketch-based method.
More precisely, this method pre-generates the simulation sketches and utilizes them to estimate the true influence~\cite{li2018influence}.
Of the sketch-based methods, the reverse reachable sketch method (RR-Sketch) is proven to be efficient.
RR-Sketch randomly selects a node $u$ and generates a simulation sketch starting from $u$ to potentially visited nodes.
Then, the visited nodes form a reverse reachable set of $u$. Repeating this process yields a collection of reverse reachable sets, which are used to estimate the influence. Several studies have extended RR-Sketch to address IBM problems.
Tong and Du~\cite{tong2019beyond} developed a hybrid sampling method that assigned a higher sampling probability to the nodes that were susceptible to misinformation. This approach enhances the optimization speed.
Manouchehri et al. \cite{manouchehri2022non} built a Non-Uniform IBM problem where users had dissimilar worthiness and introduced a sampling-based method taking the user's weight into consideration.
Yang and Li \cite{yang2024deadline} focused on the deadline-aware IBM problem and proposed an adaptive sampling approach to control the number of samples to find the optimal solution efficiently. These approaches exhibit significantly greater efficiency than MCSs for evaluations.
However, their time complexity still requires improvement as they rely on obtaining a sufficient number of samples to estimate the blocked influence accurately.

In addition, machine learning techniques have been used to construct an approximate estimator for the blocked influence.
Tong \cite{tong2020stratlearner} developed a learnable scoring model, to assess the quality of different protection strategies based on the multiple subgraphs. 
This model was then combined with the greedy algorithm to identify the optimal protector, which is called StratLearner.
Although StratLearner is faster than MCSs in influence estimation, it is still unable to provide real-time solutions.
This is because the scoring model has high time complexity due to the need of calculating the weighted sum of the distance functions for all nodes in each subgraph.
Unlike \cite{tong2020stratlearner}, our approach is analyzed to have much lower time complexity than StratLearner and can better address the IBM problems with real-time demands. 

Beyond the IBM problem, advancements have also been made in applying machine learning methods to other influence maximization (IM) problems~\cite{li2023survey,zhang2022network,chen2023touplegdd,ma2022influence}.
For instance, Li et al.~\cite{li2023influence} addressed the IM problem involving probabilistically unstable links in multi-agent systems.
Their approach utilized a graph embedding method to evaluate and select nodes based on diffusion capability, information updates, and predictions.
In another study, Tran et al.~\cite{tran2024meta} developed an IM-META method for the IM problem in networks with unknown topology by leveraging information from queries and node metadata.
They employed a Siamese neural network to learn the relationship between collected metadata and edges.
While these studies demonstrate the potential of machine learning in IM problems, they are not directly applicable to the IBM problem.
The key distinction lies in IBM's unique characteristic of involving the spread and interaction of two types of information.
This dual-information dynamic necessitates specially designed representations capable of capturing and modeling the complex interplay between different information types.

\section{Problem Definition and Notations}
\label{Problem Defination and Notations}

This section presents the information diffusion model of the social network and the formulation of the IBM problem.

\subsection{Information Diffusion Model}
\label{diffusion model}

A social network can be modeled as a directed graph $G=(V,E)$, where $V$ denotes the set of nodes and $E$ denotes the set of connections between each pair of nodes.
Suppose there are two types of information spreading in the social network: false information and true information.
Each node in the network is assigned one of three states: $F$-active (activated by the false information), $T$-active (activated by the true information), or $\Theta$-active (inactive). 

This work considers the widely used independent cascade model to simulate the influence diffusion process in IBM.
Correspondingly, there are two cascades: $C_f$ for false information and $C_t$ for true information.
Each edge $(u,v)\in E$ is assigned a propagation probability ${\rm{Pr}}[(u,v)]\in(0,1]$ to indicate the probability that node $u$ activates node $v$.
Let $S_f$ and $S_t$ denote the false-information seed set and the true-information seed set, respectively.
At step $t=0$, the diffusion of two cascades starts from $S_t$ and $S_f$.
At step $t=T(T>0)$, the $F$-active (or $T$-active) node $u$ has one chance to activate its neighbor $\Theta$-active node $v$ with the probability ${\rm{Pr}}[(u,v)]$.
When there are an $F$-active node $u_f$ and a $T$-active node $u_t$ trying to trigger their shared neighbor $\Theta$-active node $v$ simultaneously, the priority is given to $u_f$.\footnote{Our method can be adapted to the case where true information has priority.} 
Once a node is activated, its state becomes fixed, and the information continues to propagate until the states of all the node remain unchanged.
Given $G$, $S_t$, and $S_f$, the final number of $F$-active nodes is a random variable due to the inherent uncertainty of the information diffusion process.

\subsection{The IBM Problem}
Prior to introducing the IBM problem, we define non-misinformation influence and blocked influence as follows.

\begin{Definition}(Non-misinformation Influence)
	Let $\overline{F}$-active represent the state that is not $F$-active, i.e., $T$-active or $\Theta$-active.
	Given a social network $G$, a false-information seed set $S_f$, and a true-information seed set $S_t$, the non-misinformation influence $y(S_t|S_f)$ is defined as the expected number of $\overline{F}$-active nodes. That is, $y(S_t|S_f)=E[Y(S_t,\omega|S_f)]$ where the random variable $Y(S_t,\omega|S_f)$ denotes the number of $\overline{F}$-active nodes, and $\omega$ denotes the random noise representing the uncertainty regarding nodes being activated or not. 
\end{Definition}

\begin{Definition}(Blocked Influence)
	The blocked influence $f(S_t|S_f)$ is defined as the difference between the non-misinformation influence when the true-information seed set is $S_t$ and the non-misinformation influence when the true-information seed set is $\emptyset$, i.e., $f(S_t|S_f)=y(S_t|S_f)-y(\emptyset|S_f)$.
\end{Definition}

Given a social network $G$ and a false-information seed set $S_f$, the IBM problem is to select a true-information seed set $S_t$ such that the blocked influence $f(S_t|S_f)$ is maximized with the constraint on the size of $S_t$. Formally, the IBM problem is formulated as follows. 

\begin{equation}
	\label{problem 1}
	\begin{aligned}
		&\max\limits_{S_t \subset V\setminus S_f}\hspace{2ex} f(S_t|S_f)=y(S_t|S_f)-y(\emptyset|S_f)\\
		&\phantom{S t}{\rm{s.t.}}\hspace{5ex} |S_t|\leq K,
	\end{aligned}
\end{equation}
where $K\in \mathbb{Z}^+$ is the maximum size of $S_t$.

\subsection{MCSs-based Estimator}
In Problem~(\ref{problem 1}), $y(S_t|S_f)$ has no analytical form and can be estimated by Monte Carlo simulations (MCSs). 
The time complexity of MCSs is $\mathcal{O}(rm)$ where $r$ denotes the number of simulation replications and $m$ denotes the number of edges \cite{li2018influence}.
The challenge in estimating $y(S_t|S_f)$ using MCSs is the high computational burden, which can be viewed from two aspects.
Firstly, in one simulation, there is at most one chance to determine whether $u$ will activate $v$ for $\forall (u,v) \in E$.
Therefore, the worst-case time complexity for one simulation is $\mathcal{O}(m)$.
Practical social networks often consist of more than hundreds of thousands of edges, which significantly increases the time required to execute a single simulation.
Secondly, let $Y(S_t,\omega_i|S_f)$ be the $i$-th simulation sample of the number of $\overline{F}$-active nodes given random noise $\omega_i$.
Then $y(S_t|S_f)$ can be estimated by the sample mean $\overline{Y}(S_t|S_f)=\sum_{i=1}^{r}Y(S_t,\omega_i|S_f)/r$, where $r$ denotes the number of  simulation replications.
Here, we make the mild assumptions that $E[Y(S_t,\omega|S_f)]<\infty$ for $\forall S_t\subset V\setminus S_f$ and $Y(S_t,\omega_i|S_f)(i=1,...,r)$ are independent and identically distributed, which commonly hold in the simulation studies \cite{chen2020adaptive,chen2020new, chen2022multi,chen2023optimizing}.
According to the strong law of large numbers, the accurate estimation of $y(S_t|S_f)$ requires plenty of simulation replications, e.g, $r=10,000$ {\cite{kempe2003maximizing}}.

\section{Neural Influence Estimator}
{\label{Neural Network-based Surrogate Model}}

Our goal is to learn a fast approximation of MCSs, thus accelerating the optimization of IBM problems.
Specifically, we seek to build a neural influence estimator (NIE) that can 1) generalize well across different IBM problems with varying false-information seed sets, and 2) possess exceptionally low time complexity such that it can scale up to social networks involving hundreds of thousands of nodes and edges.

This section firstly formulates the learning problem and describes the structure of NIE.
Then, the time complexity of NIE is analyzed and compared with that of existing methods, followed by an NIE-based optimization method.

\begin{figure*}
	\centering
	\includegraphics[width=0.8\textwidth]{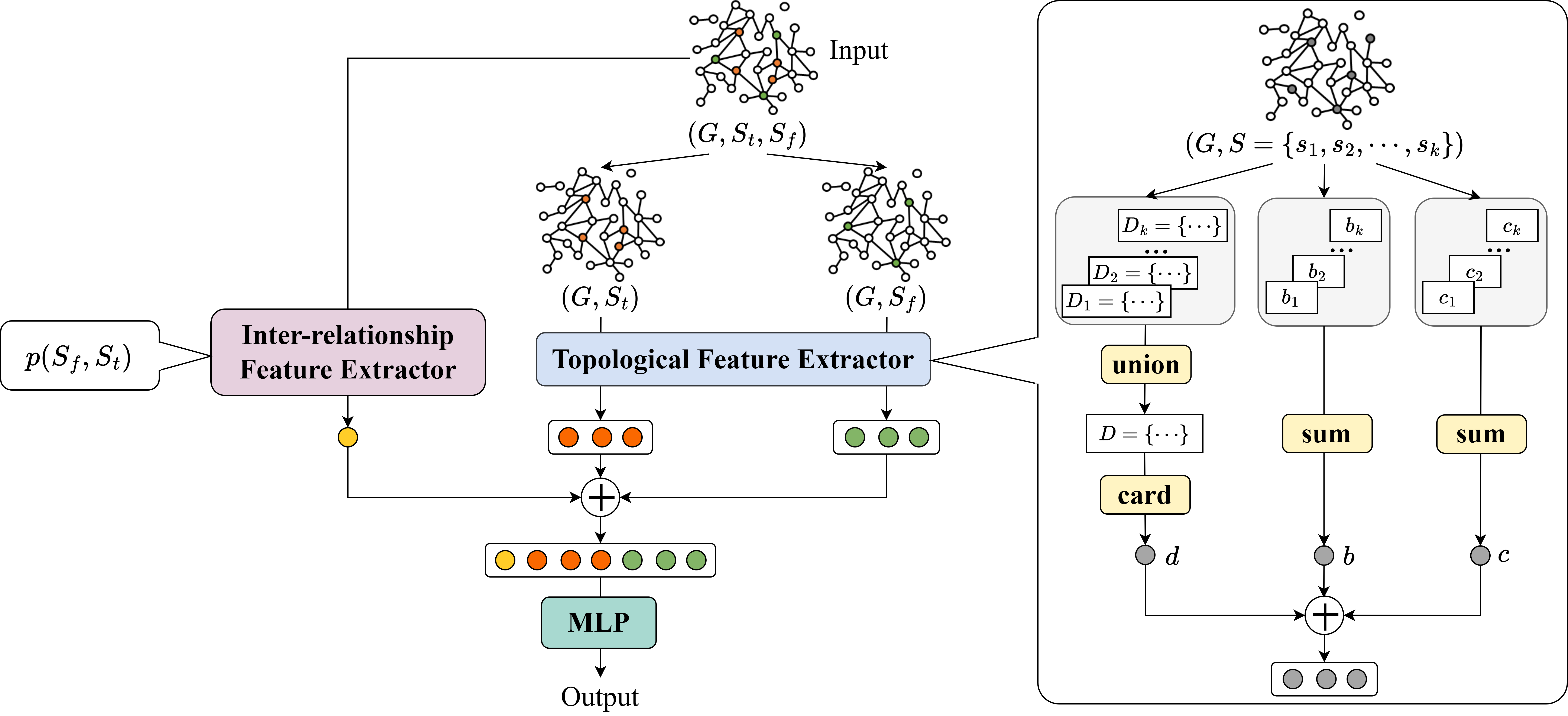} 
	\caption{The architecture of NIE. The Inter-relationship Feature Extractor includes function $p(S_f, S_t)$ that quantifies the inter-relationship between $S_f$ and $S_t$. The Topological Feature Extractor extracts the topological patterns of $S_f$ and $S_t$, including the neighborhood feature $d$, the location feature $b$, and the structure feature $c$. The MLP predicts the blocked influence based on the extracted features.}
	\label{NIE framework} 
 \vspace{-4mm}
\end{figure*}

\subsection{The Learning Problem}
In reality, social networks like Weibo and Twitter exhibit relatively stable structures, while the sources of false information tend to vary.
For example, for different types of rumors, the seed nodes can be different.
With this in mind, we seek to construct a surrogate model that can predict the blocked influence for any false-and-true information instance given a social network $G$.
Here,  a false-and-true information instance $\boldsymbol{I}$ refers to a pair of false-information seed set $S_f$ and true-information seed set $S_t$, i.e., ${\boldsymbol{I}}=(S_f,S_t)$.
Let $\hat{f}({\boldsymbol{I}})$ be the predicted blocked influence of ${\boldsymbol{I}}$ and $f_{MCSs}({\boldsymbol{I}})$ be the blocked influence estimated by MCSs.
Then, the loss function based on the mean squared error (MSE) is $Loss=(1/T)\sum_{t=1}^T[\hat{f}({\boldsymbol{I}}_t)-f_{MCSs}({\boldsymbol{I}}_t)]^2$ where $T$ is the size of training data.

\subsection{The NIE Architecture}
Neural networks have shown outstanding capabilities in modeling the nonlinear relationship between input and output, making them extensively used as surrogate models for computationally expensive modules \cite{ghafariasl2024neural,tang2024learn}.
To maintain low computational complexity, we opt for the multilayer perceptron (MLP) as the regression model because of its inherent simplicity.
However, learning the mapping from the false-and-true instance $\boldsymbol{I}$ to the blocked influence is a challenging task due to the intricate topology and inter-relationship of $S_f$ and $S_t$.
To tackle this challenge, we propose to extract informative graph features from $\boldsymbol{I}$, thereby making the subsequent prediction task tractable.
Figure~\ref{NIE framework} illustrates the architecture of NIE.

\subsubsection{Input of NIE}
The input of NIE includes a graph $G$ and a false-and-true information instance ${\boldsymbol{I}}$.
Extracting useful features from the raw input data $(G,{\boldsymbol{I}})$ poses two challenges.
\begin{itemize}
	\item The nodes in $S_f$ (or $S_t$) have tight connectivity with  the rest of nodes in $G$. $Q1:$ How to adequately extract topological patterns of $S_f$ (or $S_t$) in the whole graph?
	\item The connectivity of $S_f$ and $S_t$ greatly affects the blocked influence. $Q2:$ How to quantify the inter-relationship of the two sets in the graph?
\end{itemize}

\subsubsection{Feature Extraction}
As shown in Figure~\ref{NIE framework}, the features are extracted by two modules: the Inter-relationship Feature Extractor and the Topological Feature Extractor.
Initially, the input data $(G, S_f, S_t)$ is divided into $(G,S_f)$ and $(G,S_t)$, allowing Topological Feature Extractor to extract the topological features from $(G, S_f)$ and $(G, S_t)$ independently.
Next, the Inter-relationship Feature Extractor computes $p(S_f, S_t)$ to capture the inter-relationship between $S_f$ and $S_t$.
Finally, these features are concatenated as the input of the subsequent predictor.
The following section provides a detailed explanation of these features.

The topological patterns of $S_f$ and $S_t$ are extracted from $(G, S_f)$ and $(G, S_t)$, respectively, focusing on the aspects of neighborhood, location, and structure. This multi-perspective approach enables us to effectively capture both the local propagation potential and global bridging capacity of each node set within the network. 

\begin{itemize}
	\item \textbf{Neighborhood:} The neighborhood feature is characterized by the size of the neighbor nodes of set $S=\{s_1,s_2,...,s_k\}$, where $k$ is the size of $S$. We gather the set $D_i$ that comprises the neighbor nodes for each $s_i \in S$ and obtain the union set $D = \bigcup_{i=1}^k D_i$. The neighborhood feature $d$ is defined as the size of $D$.
	\item \textbf{Location:} The location feature reflects how central $S$ is in graph $G$ or how close $S$ is to the rest nodes in $G$. Correspondingly, the location feature $b=\sum_{i=1}^{|S|}b_i$ where $b_i=L/\sum_{l=1}^{L} d(s_i,v_l)$ is the closeness centrality of node $s_i\in S$, $L$ is the number of nodes reachable from $s_i$, $v_l$ is the reachable node from $s_i$, and $d(s_i,v_l)$ is the shortest-path distance from $s_i$ to $v_l$.
	\item \textbf{Structure:} The structure feature refers to the border structure of $S$ and is defined as the degree to which the neighbor nodes of $S$ gather into clusters. Specifically, the structure feature $c$ is the sum of all the clustering coefficient $c_i$ for $s_i\in S$ and $i=1,2,...,|S|$. Here $c_i=T(s_i) / \{2deg^{tot}(s_i) \cdot [deg^{tot}(s_i)-1] - 2deg^{\leftrightarrow}(s_i)\}$ for $s_i\in S$, where $T(s_i)$ is the number of directed triangles through node $s_i$, $deg^{tot}(s_i)$ is the sum of in-degree and out-degree of $s_i$, and $deg^{\leftrightarrow}(s_i)$ denotes the reciprocal degree of $s_i$.
\end{itemize}

\begin{algorithm*}
	\label{heuristic method}
	\caption{Approximation of ${\rm{Pr}}\{s_i\ {\rm{is}}\ F{\rm{-active}}|S_f\}$ for $\forall s_i\in S^H_f$}
	\label{algorithm 1}
	\LinesNumbered
	\KwIn {$S^H_f$, $H$.}
	\KwOut {${\rm{Pr}}\{s_i\ {\rm{is}}\ F{\rm{-active}}|S_f\}$  for $\forall s_i\in S^H_f$.}
	Set the counter $h\leftarrow 1$, and let $V_{s}$ denote the set of the parent nodes of $s$;\\
	\lFor{$\forall s_i\in S^H_f$}{calculate the propagation coefficient $pc_{s_i}\leftarrow \sum_{u\in V_{s_i}}{\rm{Pr}}[(u,s_i)]/|V_{s_i}|$}
	\lFor{$\forall s_i\in S^1_f$}{${\rm{Pr}}\{s_i\ {\rm{is}}\ F{\rm{-active}}|S_f\} \leftarrow pc_{s_i}$}
	\While{$h\leq H-1$} {
		$h\leftarrow h+1$;\\
		\lFor{$\forall s_i\in S^h_f\setminus S^{h-1}_f$}{${\rm{Pr}}\{s_i\ {\rm{is}}\ F{\rm{-active}}|S_f\} \leftarrow pc_{s_i}\cdot \left\{1-\prod_{s_j\in S^{h-1}_f\bigcap V_{s_i}}[1-{\rm{Pr}}\{s_j\ {\rm{is}}\ F{\rm{-active}}|S_f\}]\right\}$}
	}
\end{algorithm*}

Intuitively, the blocked influence $f(S_f,S_t)$ is related to the ability of $S_t$ to protect the nodes around $S_f$. Therefore, we quantify the inter-relationship between $S_f$ and $S_t$ by $p(S_f,S_t)$:
\begin{equation}
	\label{function p}
	\begin{aligned}
		&p(S_f,S_t)=\sum_{s_i\in S^H_f}[{\rm{Pr}}\{s_i\ {\rm{is}}\ F{\rm{-active}}|S_f\}\\
		&\phantom{p(S_f,S_t)= \sum_{s_i\in S^H_f}} \cdot \mathbb{1}\{d(S_t,s_i)< d(S_f,s_i)\}]
	\end{aligned}
\end{equation}
where $S^H_f$ denotes the set of nodes to which the shortest distance from $S_f$ is less than or equal to $H$-hops. 
Here, the shortest distance from a node set $S$ to a node $s_i$, denoted as $d(S, s_i)$, is $\min_{u\in S}d(u,s_i)$, where $d(u,s_i)$ is the shortest distance from $u$ to $s_i$ in graph $G$ for $u\in S$.

Specifically, $p(S_f,S_t)$ involves two components: the weight function ${\rm{Pr}}\{s_i\ {\rm{is}}\ F{\rm{-active}}|S_f\}$ and the judgment function $\mathbb{1} \{d(S_t,s_i)< d(S_f,s_i)\}$.
The weight function measures the probability that $s_i$ is activated by the false information without considering $S_t$.
A higher probability means that the node requires more attention and protection.
The judgment function evaluates the effectiveness of $S_t$ in safeguarding $s_i$.
A greater value of $p(S_f,S_t)$ indicates a stronger ability of $S_t$ to safeguard the neighbor nodes of $S_f$.

The accurate calculation of the weight function ${\rm{Pr}}\{s_i\ {\rm{is}}\ F{\rm{-active}}|S_f\}$ is intractable due to its dependence on the uncertain states of the neighbors of $s_i$.
To address this issue, we approximate ${\rm{Pr}}\{s_i\ {\rm{is}}\ F{\rm{-active}}|S_f\}$ as described in Algorithm \ref{heuristic method}.
Intuitively, a node $s_i$ is $F$-active when there exists at least one $F$-active parent node of $s_i$ and the false information can propagate to $s_i$.
Algorithm~\ref{heuristic method} first uses the average propagation probability to approximate the propagation probability of false information from parent nodes to node $s_i$ and denotes it as propagation coefficient $pc_{s_i}$ for $\forall i \in S^H_f$ (line 2).
Then, for $\forall s_i\in S^1_f$, $pc_{s_i}$ is directly used to approximate ${\rm{Pr}}\{s_i\ {\rm{is}}\ F{\rm{-active}}|S_f\}$ (line 3);
while for $\forall s_i\in S^h_f\setminus S^{h-1}_f(2\leq h\leq H)$,  ${\rm{Pr}}\{s_i\ {\rm{is}}\ F{\rm{-active}}|S_f\}$ is determined by the product of the propagation coefficient and the probability that there exists at least one parent node being $F$-active (lines 4-7).

To calculate the judgement function $\mathbb{1} \{d(S_t,s_i)< d(S_f,s_i)\}$, the shortest distances between paired nodes in graph $G$ can be pre-computed and collected offline. During the online optimization process, $d(S_t,s_i)$ (or $d(S_f,s_i)$) is determined by minimizing $d(u,s_i)$ for $u\in S_t$ (or $u\in S_f$). Subsequently, $\mathbb{1} \{d(S_t,s_i)< d(S_f,s_i)\}$ is derived through the comparison between $d(S_t,s_i)$ and $d(S_f,s_i)$.

\subsubsection{MLP Layer} The extracted features are fed into MLP, and the output is the predicted blocked influence $\hat{f}({\boldsymbol{I}})$. For MLP, the hidden layers are fully connected followed by a rectified linear unit (ReLU) activation function. Early stopping is used to prevent overfitting.

\subsection{Complexity Analysis and Comparison}
\label{complexity analysis}
This section analyzes the time complexity of NIE and compares it with other influence estimators.
For brevity, we denote $n=|V|$, $m=|E|$, $k_f=|S_f|$, and $k_t=|S_t|$.

\subsubsection{Complexity Analysis for NIE}
The main components in NIE consist of the Topological Feature Extractor, the Inter-relationship Feature Extractor, and MLP.
Among them, the time complexity of MLP is independent of network scale (i.e., the number of nodes and edges).
In practice, the computational time of MLP is almost negligible, compared to the computational time of the other two components.
Hence, we only analyze the time complexities of the Topological Feature Extractor and the Inter-relationship Feature Extractor.

The time complexity of the Topological Feature Extractor is determined by the \textbf{union} operation.\footnote{Given a graph, we can pre-compute and collect data about node neighbors, closeness centrality, and clustering coefficient offline. These stored data are then used in online optimization. Therefore, we only focus on the time complexity of the union operation in the Topological Feature Extractor.}
The time complexities of the Topological Feature Extractor for $(G,S_f)$ and the Topological Feature Extractor for $(G,S_t)$ are $\mathcal{O}(k_f\cdot\min(m,n))$ and $\mathcal{O}(k_t\cdot\min(m,n))$ respectively, because the Hash table can be used to obtain $D$ and the number of nodes connected to $S_f$ (or $S_t$) is bounded by $m$ and $n$.
Hence, the overall time complexity of the Topological Feature Extractor is  $\mathcal{O}((k_f+k_t) \cdot\min(m,n))$.

To compute $p(S_f,S_t)$ for the Inter-relationship Feature Extractor, we firstly construct set $S_f^H$. Let $S_f^H=\emptyset$.
The seed nodes in $S_f$ are sequentially chosen as the initial point $x_0$ to spread the influence. The breadth-first search is performed to visit the node $s_i$ whose distance to $x_0$ is less than or equal to $H$. If $s_i\notin S_f^H$, add $s_i$ into $S_f^H$. The above process is repeated until all solutions in $S_f$ are traversed. The size of $S_f^H$ is bounded by $m$ and $n$. Therefore, the worst-case time complexity for constructing $S_f^H$ is $\mathcal{O}(\min(m,n))$. Secondly, for $\forall s_i\in S_f^H$, $d(S_f,s_i)$ and $d(S_t,s_i)$ are selected by the BFPRT algorithm \cite{blum1973time}. Their worst-case time complexities are $\mathcal{O}(k_f\cdot\min(m,n))$ and $\mathcal{O}(k_t\cdot\min(m,n))$, respectively, because $|S_f^H|$ is bounded by $m$ and $n$ and the worst-case time complexity for selecting $d(S_f,s_i)$ (or $d(S_t,s_i)$) is $\mathcal{O}(k_f)$ (or $\mathcal{O}(k_t)$). Lastly, the ${\rm{Pr}}(s_i\ {\rm{is}}\ F{\rm{-active}}|S_f)$ is computed for $\forall s_i\in S^H_f$. As shown in Algorithm \ref{algorithm 1}, the false information propagates from the nodes in layer $h-1$ to the nodes in layer $h$ ($h=2,\dots,H$).
If there exists an edge from $s_j$ ($s_j\in S_f^{h-1}$) to $s_i$ ($s_i\in S_f^h$), $1-{\rm{Pr}}(s_j\ {\rm{is}}\ F{\rm{-active}}|S_f)$ is calculated in Step 5.
The number of calculations is bounded by $m$ and $n$ because the number of the edges between all the adjacent layers is less than or equal to $m$ and $|S_f^H|$ is bounded by $m$ and $n$.
Therefore, the worst-case time complexity of Algorithm~\ref{algorithm 1} is $\mathcal{O}(\min(m,n))$.
Consequently, the worst-case time complexity of computing $p(S_f,S_t)$ is $\mathcal{O}((k_f+k_t)\cdot\min (m,n))$.
Combining the complexity of the Topological Feature Extractor, the overall complexity of NIE is $\mathcal{O}((k_f+k_t)\cdot\min (m,n))$.

\begin{table}[tbp]
	\caption{Comparison in terms of time complexities of different influence estimators.}
	\label{complexity}
	\centering
	\begin{tabular}{p{1cm}m{3cm}<{\centering}p{3.7cm}}
		\toprule
		\textbf{Estimator}   & \textbf{Time Complexity}  & \multicolumn{1}{c}{\textbf{Explanation}} \\ \midrule
		\multirow{5}{*}{NIE} & \multirow{5}{*}{$\mathcal{O}((k_f+k_t)\cdot\min (m,n))$} & $n,m,k_f,k_t$ are the node number, edge number, size of the false-information seed set $S_f$, and size of the true-information seed set $S_t$, respectively.\\
		\midrule
		\multirow{2}{*}{MCSs} & \multirow{2}{*}{$\mathcal{O}(r\cdot m)$} & $r$ is the number of simulation replications.\\
		\midrule
		HMP* &  $\mathcal{O}(l\cdot k_t\cdot \min(m,n))$ & $l$ is the number of $R$-samples.\\
		\midrule
		StratLearner* & $\mathcal{O}(t \cdot (k_f+k_t)\cdot n)$ & $t$ is the number of subgraphs.\\
		\bottomrule
	\end{tabular}
	\vspace{-4mm}
\end{table}

\subsubsection{Complexity Comparison of Different Influence Estimators}
We compare NIE with three existing influence estimators used by the state-of-the-art IBM methods.
\textbf{MCSs} is the baseline approach to estimating the blocked influence.
\textbf{HMP*} represents the estimator used in the sketch-based method HMP {\cite{tong2019beyond}}.
{\textbf{StratLearner*}} is a learning-based surrogate model that estimates influence in the method StratLearner {\cite{tong2020stratlearner}}.
Table~\ref{complexity} presents the time complexities.
For detailed analyses for HMP* and StratLearner*, please refer to the Supplement.

For MCSs, the number of simulation replications $r$ should be large enough, empirically $r=10,000$~\cite{kempe2003maximizing}.
For StratLearner*, it is recommended that the number of subgraphs $t$ should be at least 400 for accurate estimations~\cite{tong2020stratlearner}.
For HMP*, $l$ is the number of $R$-samples, which should be set sufficiently large as recommended in~\cite{tong2019beyond}.
In practice, it holds that $k_t+k_f \ll r$ and $k_t+k_f \ll l$.
Hence,  the time complexity of NIE is much lower than that of MCSs, HMP*, and StratLearner*.

\subsection{NIE-based Optimization Method}
\label{NIE-based greedy algorithm}
The NIE can be integrated into any MCSs-based optimization method for solving IBM problems.
In this work, we integrate NIE into the cost-effective lazy forward selection algorithm (CELF), an advanced variant of the greedy algorithm~\cite{leskovec2007cost}.
The resultant method, called NIE-CELF, is illustrated in Algorithm~\ref{greedy method}.
It is worth mentioning that NIE-CELF can be further combined with heuristics to reduce the number of evaluations and achieve even higher efficiency {\cite{yan2019minimizing}}.

\begin{algorithm}[tbp]
	\label{greedy method}
	\caption{NIE-CELF}
	\LinesNumbered
	\KwIn {$K$, $S_f$, $G$}
	\KwOut {$S_t$}
	Set the counter $k\leftarrow 0$ and $S_t\leftarrow \emptyset$;\\
	\While{$k\leq K-1$} {
		$k\leftarrow k+1$;\\
		$v^*\leftarrow LazyForward(\hat{f}({\boldsymbol{I}}), S_f, S_t, G)$~\cite{leskovec2007cost};\\
		$S_t \leftarrow S_t\cup\{v^*\}$.
	} 
\end{algorithm}

\section{Experimental Studies}
\label{Experimental Studies}

The experiments aim to assess the efficiency and scalability of NIE.
We select five social networks with varying scales, generate 25 IBM problems, and compare NIE-CELF with five state-of-the-art methods.
Specifically, two sets of experiments are conducted.
In the first experiment, the performance of different methods is compared in terms of the runtime required to achieve a predefined optimization quality.
In the second experiment, the methods are compared from the perspective of optimization quality obtained within a time limit.
In addition, the feature importance of NIE, as well as the settings of $H$ and the hyperparameters in MLP, are also discussed.

\subsection{Experimental Setup}
\subsubsection{Networks and Test Problems}

We consider five networks with different scales, structures, and average degrees from~\cite{tong2020stratlearner} and the SNAP datasets.
Table~\ref{datasets} summarizes their statistics.
\textbf{Power-law graph} simulates the common power-law distributed network.
\textbf{Email-Eu-core} is built based on email data from a large European research institution to describe the email network across 42 different departments.
\textbf{P2p-Gnutella08} and \textbf{p2p-Gnutella24} are established based on the snapshot data from the Gnutella peer-to-peer file sharing network to describe the connections between different hosts in the network.
\textbf{Web-Stanford}, sourced from the Stanford University website, describes the hyperlink relationship between different pages.

\begin{table}[t]
	\caption{Network statistics.} 
	\label{datasets}
	\centering
	\setlength{\tabcolsep}{2.8mm}{
		\begin{tabular}{cccc}
			\toprule
			\textbf{Network} & \textbf{\# Nodes} & \textbf{\# Edges} & \textbf{\begin{tabular}[c]{@{}c@{}}Average Degree\end{tabular}} \\ 
			\midrule
			power-law graph  & 768               & 1,532             & 1.99                                                               \\
			email-Eu-core    & 1,005             & 25,571            & 25.44                                                              \\
			p2p-Gnutella08   & 6,301             & 20,777            & 3.30                                                               \\
			p2p-Gnutella24   & 26,518            & 65,369            & 2.47                                                               \\
			web-Stanford     & 281,903           & 2,312,497         & 8.20                                                               \\ 
			\bottomrule
	\end{tabular}}
	\vspace{-4mm}
\end{table}

For each network, we generate five IBM problems, meaning that there are 25 test problems in total.
Specifically, each problem is generated in the following way.
We rank all the nodes according to their out-degrees, add the top $\rho$ nodes into a high-impact set $V^*$, and then randomly select $S_f$ from $V^*$, i.e., each $S_f$ corresponds to a test problem.
The size of $S_f$ is sampled from a power-law distribution with the shape parameter 9 and the scale parameter 10.
The above selection rule for $S_f$ is reasonable in practice because the false information typically targets nodes with high impact to achieve wider dissemination. 
We set $\rho=1\%$ for the web-Stanford network and $\rho=10\%$ for other networks.
For each test problem, the maximum size of $S_t$, i.e., $K$ in Problem~(\ref{problem 1}), equals to $|S_f|$.
For the information diffusion model (see Section~\ref{diffusion model}), we adopt the commonly used propagation probability $P[(u,v)]=1/d(v)$ where $d(v)$ is the in-degree of node $v$ \cite{tong2019beyond}.

\subsubsection{Compared Methods}
NIE-CELF is compared with five state-of-the-art methods: MCSs-CELF, HMP*-CELF, StratLearner*-CELF, CMIA-O, and ACO-GE. MCSs-CELF, HMP*-CELF, and StratLearner*-CELF respectively embody the simulation-based, sketch-based, and machine learning-based solution evaluation approaches for IBM problems. These methods integrate influence estimators MCSs, HMP*, and StratLearner* with the same optimization algorithm, CELF.\footnote{In the original publications of HMP~\cite{tong2019beyond} and StratLearner~\cite{tong2020stratlearner}, these methods are implemented using the greedy algorithm.
Compared to the greedy algorithm, CELF achieves higher efficiency without compromising optimization quality. Therefore, we replace the greedy algorithm with CELF.} CMIA-O~\cite{wu2017scalable} is a heuristic method tailored for IBM problems, enhancing efficiency by constraining influence propagation paths. It exhibits superior speed compared to the MCSs-greedy algorithm. Different from the above search-based methods, ACO-GE~\cite{chen2022graph} is a recent rank-based approach that sorts nodes and selects the highest-ranked ones.\footnote{ACO-GE first employs graph embedding to discern node relationships and then utilizes the ant colony behavior to simulate negative influence propagation, identifying the candidate nodes. Finally, it chooses candidate nodes with significant impacts.}

\subsubsection{Performance Metrics}
The concerned performance metrics are optimization speed and optimization quality.
We measure the optimization speed using the runtime, denoted as $T$. {Specifically, the runtime $T$ represents the actual time required for our reference machine to execute an IBM algorithm.
The optimization quality is measured by the blocked influence $f({S_t^*}|S_f)$ as defined in Definition 2, where $S_t^*$ is the solution found by the IBM algorithm. Because $f({S_t^*}|S_f)$ cannot be obtained accurately, 
we utilize $\hat{f}_{MCSs}(\hat{S_t^*}|S_f)$ to estimate the value of $f({S_t^*}|S_f)$. $\hat{f}_{MCSs}(\hat{S_t^*}|S_f)$ is the blocked influence evaluated by MCSs with $\hat{S_t^*}$ being the final solution found by the IBM algorithm.}

\subsubsection{Experimental Protocol} \label{experiment setting}

\begin{table}[tbp]
\caption{Parameter settings for the compared methods.}
\label{parameter settings}
\centering
\setlength{\tabcolsep}{3mm}{
	\begin{tabular}{ccc}
		\toprule
		\textbf{Method}             & \textbf{Parameter Description} & \textbf{Value} \\ 
		\midrule
		MCSs-CELF                      & Simulation replications        & 10,000       \\
		\cmidrule(lr){2-3}
		HMP*-CELF                         & Number of $R$-sampled     & 10,000     \\
		\cmidrule(lr){2-3}
		\multirow{2}{*}{StratLearner*-CELF} & Training set size                  & 192        \\
		& Number of subgraphs            & 400           \\ 
		\cmidrule(lr){2-3}
		CMIA-O                          & Influence threshold            & 0.01        \\
		\cmidrule(lr){2-3}
          \multirow{3}{*}{ACO-GE} & Maximum number of iterations                  & 200        \\& Length of the ant selection path            & 10           \\ & Number of ants            & $3*|S_f|$           \\ 
		\bottomrule
\end{tabular}}
\vspace{-4mm}
\end{table}

\begin{table*}[tbp]
	\centering
	\caption{The runtime (in seconds) required by the methods to achieve the target optimization quality. The lower, the better.}
	\label{runtime}
	\setlength{\tabcolsep}{2mm}{
		\begin{threeparttable}{
				\begin{tabular}{cccccccccc}
					\toprule
					\multirow{3}{*}{\textbf{Network}} &\multirow{3}{*}{{$\bm{|V|}$}} &\multirow{3}{*}{{$\bm{|E|}$}}&\multirow{3}{*}{\textbf{Problem}} & \multicolumn{6}{c}{\textbf{Method}}       \\ \cmidrule{5-10} 
					& && & NIE-CELF & MCSs-CELF   & HMP*-CELF     & StratLearner*-CELF & CMIA-O & ACO-GE\\ 
					\midrule
					\multirow{5}{*}{power-law graph} & \multirow{5}{*}{768} & \multirow{5}{*}{1,532}  & 1                      & \textbf{0.27}      & 184.18     & 7.49   & 117.35   & 0.36  & \underline{1.32} \\
					&& & 2                    & 0.41      & 117.23    & 6.77   & 101.69  & \textbf{0.39} & \underline{1.11} \\
					& &&3                     & \textbf{0.19}      & 128.48     & 4.47    & 65.02    & 0.26 & \underline{1.03}\\
					& &&4                     & \textbf{0.18}      & 232.55    & 14.59   & 193.08   & 1.29 & \underline{1.43} \\
					&&&5                     & \textbf{0.23}      & 294.01    & 16.87   & 274.50   & 1.33 & \underline{1.43}\\ 
					\midrule[.02em]
					Average speedup && && & 842.91 & 44.44 & 658.20 & 3.32 & 5.44 \\
					Comparison result  & & && & w &w &w &d &  w\\
					\midrule
					\multirow{5}{*}{email-Eu-core} & \multirow{5}{*}{1,005} & \multirow{5}{*}{25,571} & 1 & \textbf{0.78}      & 9943.50 & 51.27   & 166.74   & \underline{86.43} & \underline{17.92} \\
					&&&2                        & \textbf{0.60}      & 16092.73  & 50.74   & 253.14   & \underline{77.16} & \underline{16.96} \\
					&&&3                        & \textbf{1.06}      & 3576.73   & 36.25   & 189.28  & \underline{93.67} & \underline{16.36}  \\
					&&&4                        & \textbf{0.72}      & 4839.09  & 33.82   & 192.95   & \underline{109.01}& \underline{16.08} \\
					&&&5                        & \textbf{0.65}      & 6458.17  & 36.34  & 163.83  & \underline{91.35} & \underline{16.29}  \\ 
					
					\midrule[.02em]
					Average speedup  & &&& & 11920.03    & 57.48  & 266.85  & 123.94 & 22.81 \\
					Comparison result && & & & w& w&w &w & w\\
					\midrule
					
					\multirow{5}{*}{p2p-Gnutella08} & \multirow{5}{*}{6,301} & \multirow{5}{*}{20,777} &1                       & \textbf{0.31}      & 8370.16 & 175.69  & \underline{172800.00}    & \underline{158.81}  & \underline{10.16} \\
					&&&2                       & \textbf{0.25}      & 2629.84   & 132.94  & \underline{172800.00}      & 112.22& \underline{10.24} \\
					&&&3                       & \textbf{0.25}      & 5815.10  & 194.24  & \underline{172800.00}       & 234.29 & \underline{10.32}\\
					&&&4                       & \textbf{0.24}      & 7308.41  & 185.82  & \underline{172800.00}      & 169.95 & \underline{10.04} \\
					&&&5                       & \textbf{0.26}      & 3178.32    & 82.34  & \underline{172800.00}   & \underline{43.81} & \underline{10.06} \\ 
					
					\midrule[.02em]
					Average speedup  & &&& & 20691.26 & 593.28  &664886.95  & 554.99 & 39.11 \\
					Comparison result  & &&& &w &w&w &w & w \\
					\midrule
					
					\multirow{5}{*}{p2p-Gnutella24} & \multirow{5}{*}{26,518} & \multirow{5}{*}{65,369}&1                        & \textbf{0.72}      & \underline{172800.00}          & 220.60  & \underline{172800.00}      & \underline{44.47}    & \underline{33.77}     \\
					&&&2                       & \textbf{0.65}      & \underline{172800.00}         & 746.72 & \underline{172800.00}      & 237.38     & \underline{34.59}    \\
					&&&3                       & \textbf{0.64}      & \underline{172800.00}        & 473.68 & \underline{172800.00}         & 104.56   & \underline{33.94}   \\
					&&&4                       & \textbf{0.66}      & \underline{172800.00}         & 346.81 & \underline{172800.00}       & \underline{72.45}        & \underline{34.14} \\
					&&&5                       & \textbf{0.65}      & \underline{172800.00}         & 367.80 & \underline{172800.00}      & 209.82   & \underline{34.20}     \\ 
					
					\midrule[.02em]
					Average speedup  & &&& & 260702.10 & 657.33  &260702.10& 204.58 & 51.50\\
					Comparison result  &&& & & w&w &w &w & w \\
					\midrule
					
					\multirow{5}{*}{web-Stanford} & \multirow{5}{*}{281,903} & \multirow{5}{*}{2,312,497}&1                         & \textbf{26.11}     & \underline{172800.00}         & 95.27  & \underline{172800.00}   & \underline{127014.48}    & \underline{1084.54}      \\
					&&&2                         & \textbf{29.69}     & \underline{172800.00}        & 253.54  & \underline{172800.00}      & \underline{128309.73}   & \underline{1082.29}    \\
					&&&3                         & \textbf{23.03}     & \underline{172800.00}         & 108.13  & \underline{172800.00}     & \underline{5091.59}   & \underline{1119.44}     \\
					&&&4                         & \textbf{26.27}     & \underline{172800.00}        & 218.33  & \underline{172800.00}     & \underline{95039.35}    & \underline{1111.63}    \\
					&&&5                         & \textbf{21.08}     & \underline{172800.00}          & 108.58  & \underline{172800.00}      & \underline{7540.60}    & \underline{1078.91}   \\ 
					
					\midrule[.02em]
					Average speedup  &&& & & 6943.35& 6.07  & 6943.35& 2676.57 & 44.02  \\
					Comparison result  &&& & &w &w & w&w & w \\
					\bottomrule
			\end{tabular}}
			\begin{tablenotes}
				\footnotesize
				\item[a] For each test problem, the best runtime is indicated in bold.
				\item[b] The underline ``\underline{\ }'' means that the method fails to reach the target optimization quality when it is terminated due to either reaching the maximum iterations or using up the time budget of 48 hours.
				\item[c] ``Average speedup" indicates the average runtime ratio between NIE-CELF and the compared method, across all the five test problems on the network.
				\item[d] ``Comparison result" indicates the result of the Mann-Whitney U test with a significance level of 0.05. 
				``w'' (or ``l'') means that the runtime of NIE-CELF is significantly better (or worse) than that of the compared method. ``d'' means that the runtime difference is not significant.
			\end{tablenotes}
	\end{threeparttable}}
\vspace{-4mm}
\end{table*}

The parameters of the compared methods are set as recommended in the literature and are summarized in Table~\ref{parameter settings}.
For NIE-CELF, we conduct preliminary experiments to determine $H$ and the hyperparameters of MLP. 
Specifically,  $H$ is set to 3 for the web-Stanford network and 2 for other networks.
For MLP, two hidden layers are applied and the dimension of each hidden layer is 128; batch size and learning rate are set to 512 and 0.05, respectively.
To train NIE, for each network, we generate 100,000 training examples.
Each example contains a false-and-true information instance ${\boldsymbol{I}}=(S_f,S_t)$ and its corresponding blocked influence estimated by 1,000 replications of MCSs.
Here, $S_f$ is randomly selected as previously described and $S_t$ is randomly selected from $V\setminus S_f$ with $|S_t|\leq |S_f|$ because we set the maximum size of $S_t$ as $|S_f|$. 

Since MCSs-CELF, HMP*-CELF and ACO-GE contain randomized components, their results are obtained by executing each method for 5 independent runs.
For all the compared methods, $T$ and $\hat{f}_{MCSs}(\hat{S_t^*}|S_f)$ in each iteration are recorded.
The termination condition is either reaching the maximal number of iterations, or exceeding the runtime limit of 48 hours.
To make a fair comparison, all methods are implemented in Python and run on the same computing platform, i.e.,  Intel Xeon Gold 6336Y CPU with 3.6GHz and 256GB of memory.

\begin{table*}[tbp]
	\centering
	\caption{The optimization quality (blocked influence) achieved by the methods with a time budget of one minute. The higher, the better.}
	\label{quality}
	\setlength{\tabcolsep}{2mm}{
		\begin{threeparttable}{
				\begin{tabular}{cccccccccc}
					\toprule
					\multirow{3}{*}{\textbf{Network}} &\multirow{3}{*}{{$\bm{|V|}$}} &\multirow{3}{*}{{$\bm{|E|}$}}&\multirow{3}{*}{\textbf{Problem}} & \multicolumn{5}{c}{\textbf{Method}}       \\ \cmidrule{5-10} 
					&&& & NIE-CELF & MCSs-CELF  & HMP*-CELF     & StratLearner*-CELF & CMIA-O  & ACO-GE \\ 
					\midrule
					\multirow{5}{*}{power-law graph} & \multirow{5}{*}{768} & \multirow{5}{*}{1,532} 
					& 1                     & 44.39     & -     & \textbf{56.34}   & -   & 53.83    & 19.80   \\
					&&& 2                    & 51.85     & -     & \textbf{75.33}   & -  & 73.69   & 16.10  \\
					&&&3                     & 40.75     & 28.19   & \textbf{45.02}   & -    & 44.31   & 12.87  \\
					&&&4                     & 74.83     & -    & \textbf{106.98}  & -  & 96.84  & 17.17 \\
					&&&5                     & 84.65      & -      & \textbf{107.48}  & -    & 95.00  & 24.82\\ 
					\midrule[0.02em]
					Average quality ratio && && & - & 0.77 & - & 0.82 & 3.28 \\
					Comparison result  &&& & & - & d & - & d & w\\
					\midrule
					\multirow{5}{*}{email-Eu-core} & \multirow{5}{*}{1,005} & \multirow{5}{*}{25,571} 
					& 1                       & \textbf{51.67}      & -     & 49.71 & -  & -    & 28.90  \\
					&&&2                        & 55.50      & -      & \textbf{55.72}  & - & -  & 25.18\\
					&&&3                        & 38.89     & -    & \textbf{49.43}  & - & -  & 19.68 \\
					&&&4                        & 38.05     & -   & \textbf{52.60}   & - & -  & 21.82 \\
					&&&5                        & 41.05      & -    & \textbf{49.29} & - & -  & 19.79\\ 
					\midrule[0.02em]
					Average quality ratio && && & - & 0.88 & - & - & 1.96 \\
					Comparison result  & &&& & - & d & -  &- & w\\
					\midrule
					
					\multirow{5}{*}{p2p-Gnutella08} & \multirow{5}{*}{6,301} & \multirow{5}{*}{20,777} 
					&1                       & \textbf{210.37}      & -  & -  & -      & -        & 71.34  \\
					&&&2                       & \textbf{107.26}     & -   & -  & -      & -   & 74.33 \\
					&&&3                       & \textbf{352.04}     & - & - & -     & -     & 134.09 \\
					&&&4                       & \textbf{300.95}     & -  & -  & -    & -    & 121.10 \\
					&&&5                       & \textbf{86.34}      & -  & -  & -    & 63.01       & 39.03 \\ 
					\midrule[0.02em]
					Average quality ratio && && & - & - & - & - & 2.34 \\
					Comparison result  & &&& & -  & - &- &- & d\\
					\midrule
					
					\multirow{5}{*}{p2p-Gnutella24} & \multirow{5}{*}{26,518} & \multirow{5}{*}{65,369} 
					&1                        & \textbf{123.02}      & -         & -  & -     & 47.55  & 52.68 \\
					&&&2                       & \textbf{385.96}     & -        & - & -    & -   & 163.62  \\
					&&&3                       & \textbf{340.15}      & -         & - & -    & 92.59 &  189.58 \\
					&&&4                     & \textbf{285.00}     & -         & - & -     & 196.06 & 138.40 \\
					&&&5                       & \textbf{323.32}     & -       & - & -      & -  &  77.61\\ 
					\midrule[0.02em]
					Average quality ratio && && & - & - & - & - & 2.54 \\
					Comparison result  & &&& & -&- &- &- & d \\
					\midrule
					
					\multirow{5}{*}{web-Stanford} & \multirow{5}{*}{281,903} & \multirow{5}{*}{2,312,497} 
					&1                         & \textbf{122.82}    & -        & -  & -     & -  & -      \\
					&&&2                         & \textbf{183.60}    & -         & -  & -    & -  & -      \\
					&&&3                         & \textbf{112.22}     & -         & -  & -    & -  &  -    \\
					&&&4                         & \textbf{229.60}     & -        & -  & -    & -  &  -     \\
					&&&5                         & \textbf{82.06}    & -        & -  & -    & -  &  -     \\ 
					\midrule[0.02em]
					Average quality ratio && && & - & - & - & - & - \\
					Comparison result  & &&& &-&- & - &- &- \\
					\bottomrule
			\end{tabular}}
			\begin{tablenotes}
				\footnotesize
				\item[a] For each test problem, the best optimization quality is indicated in bold.
				\item[b] ``-" means that the method fails to find a solution with a time budget of one minute.
				\item[c] ``Average quality ratio" indicates the average quality ratio between NIE-CELF and the compared method across all the five test problems on the network.
				\item[d] ``Comparison result" indicates the result of the Mann-Whitney U test with a significance level of 0.05. 
				``w'' (or ``l'') means that the optimization quality of NIE-CELF is significantly better (or worse) than that of the compared method. ``d'' means that the quality difference is not significant.
			\end{tablenotes}
	\end{threeparttable}}
	\vspace{-4mm}
\end{table*}

\begin{figure*}[tbp]
	\centering
	\subfloat{\includegraphics[width=1.7in]{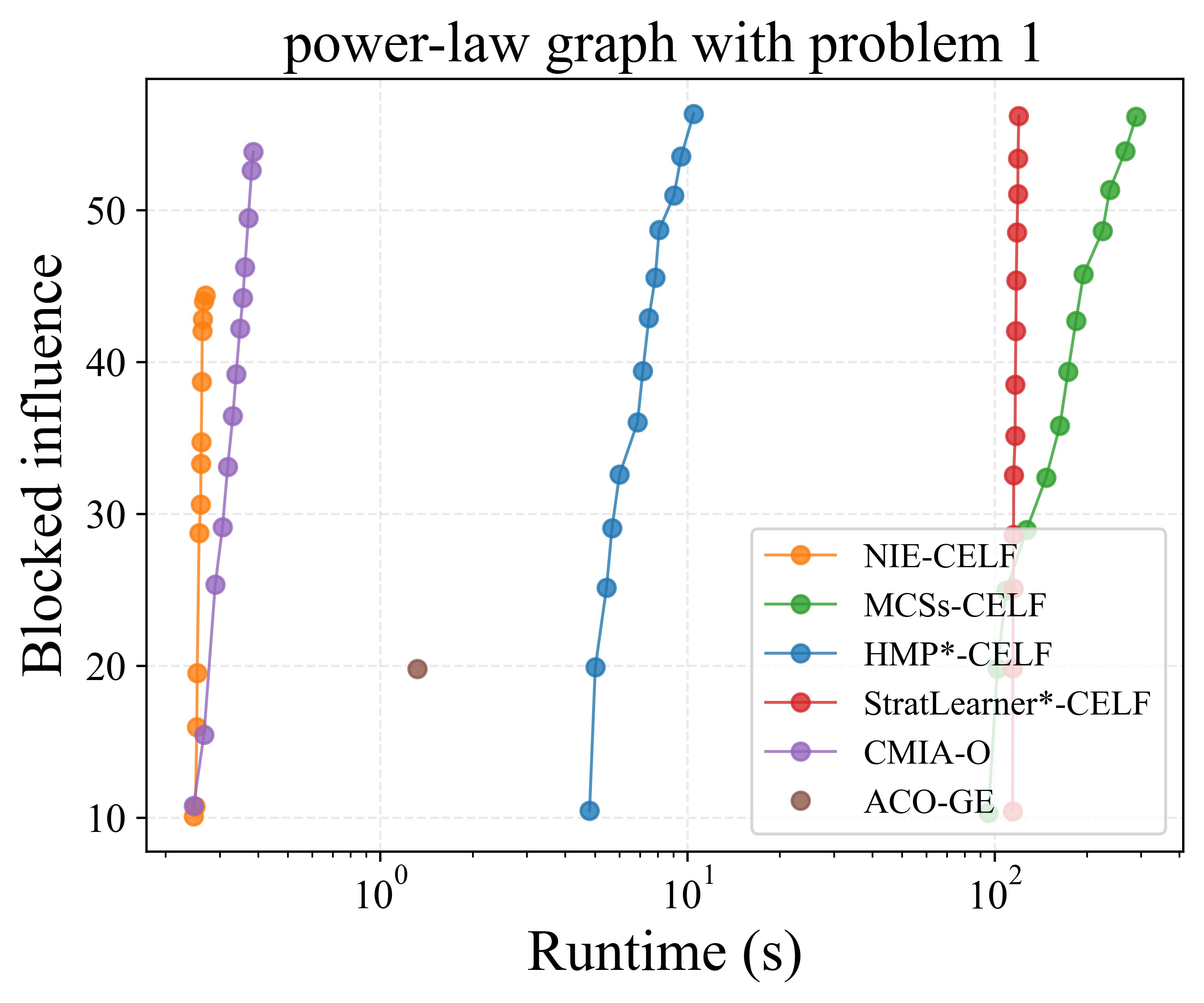}}
	\hfil
	\subfloat{\includegraphics[width=1.7in]{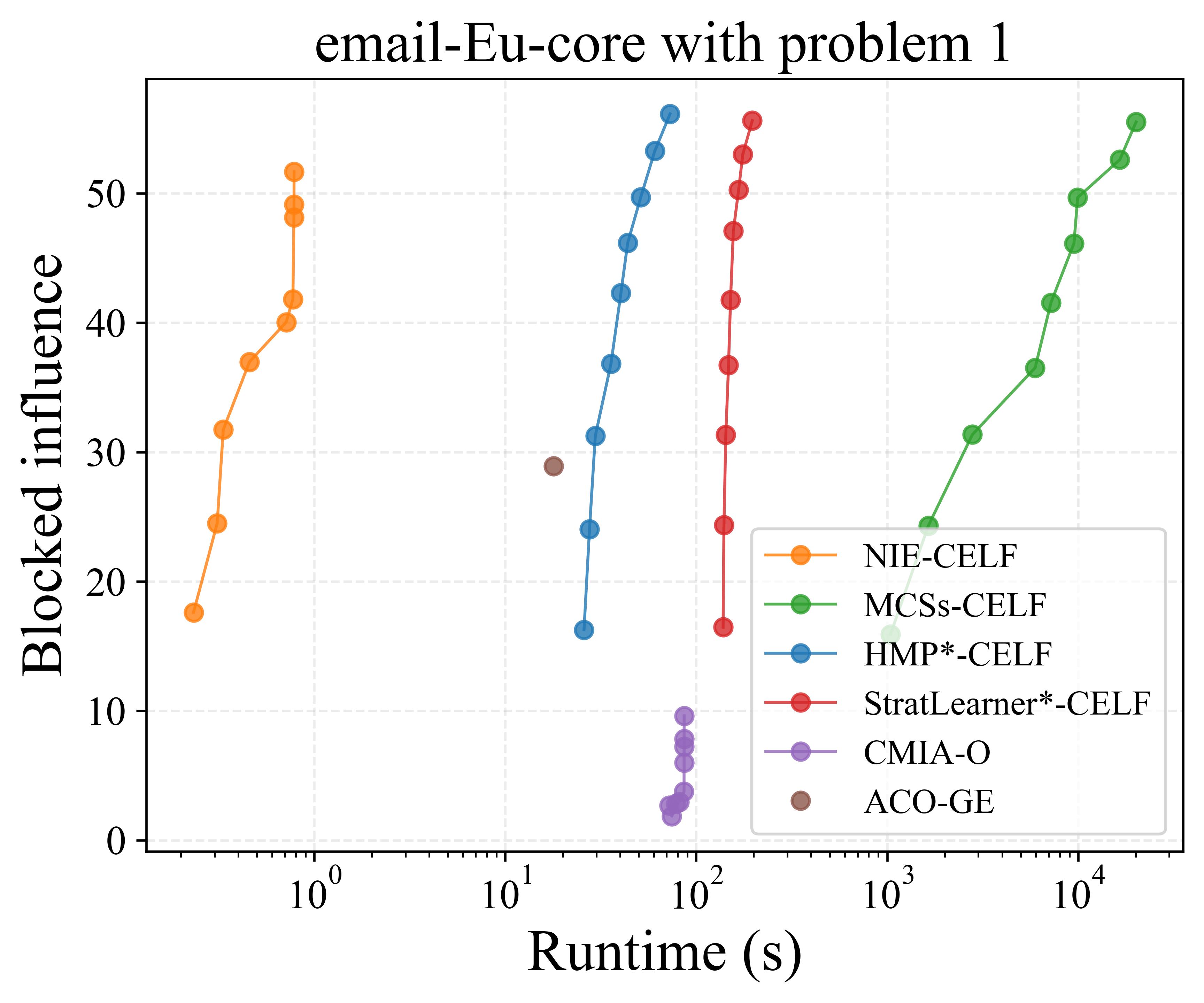}}
	\hfil
	\subfloat{\includegraphics[width=1.7in]{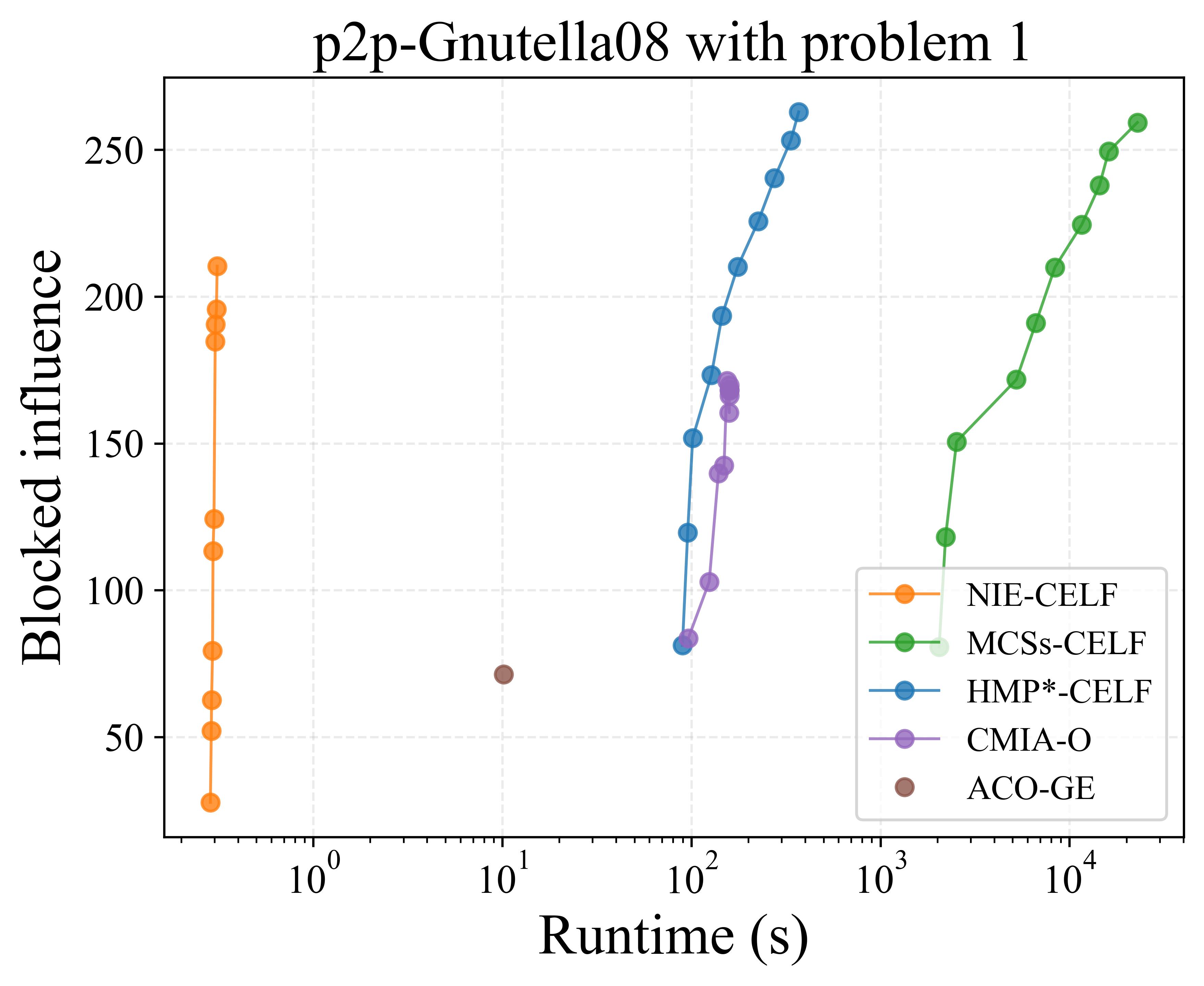}}
	\hfil \\ \vspace{-2mm}
	\subfloat{\includegraphics[width=1.7in]{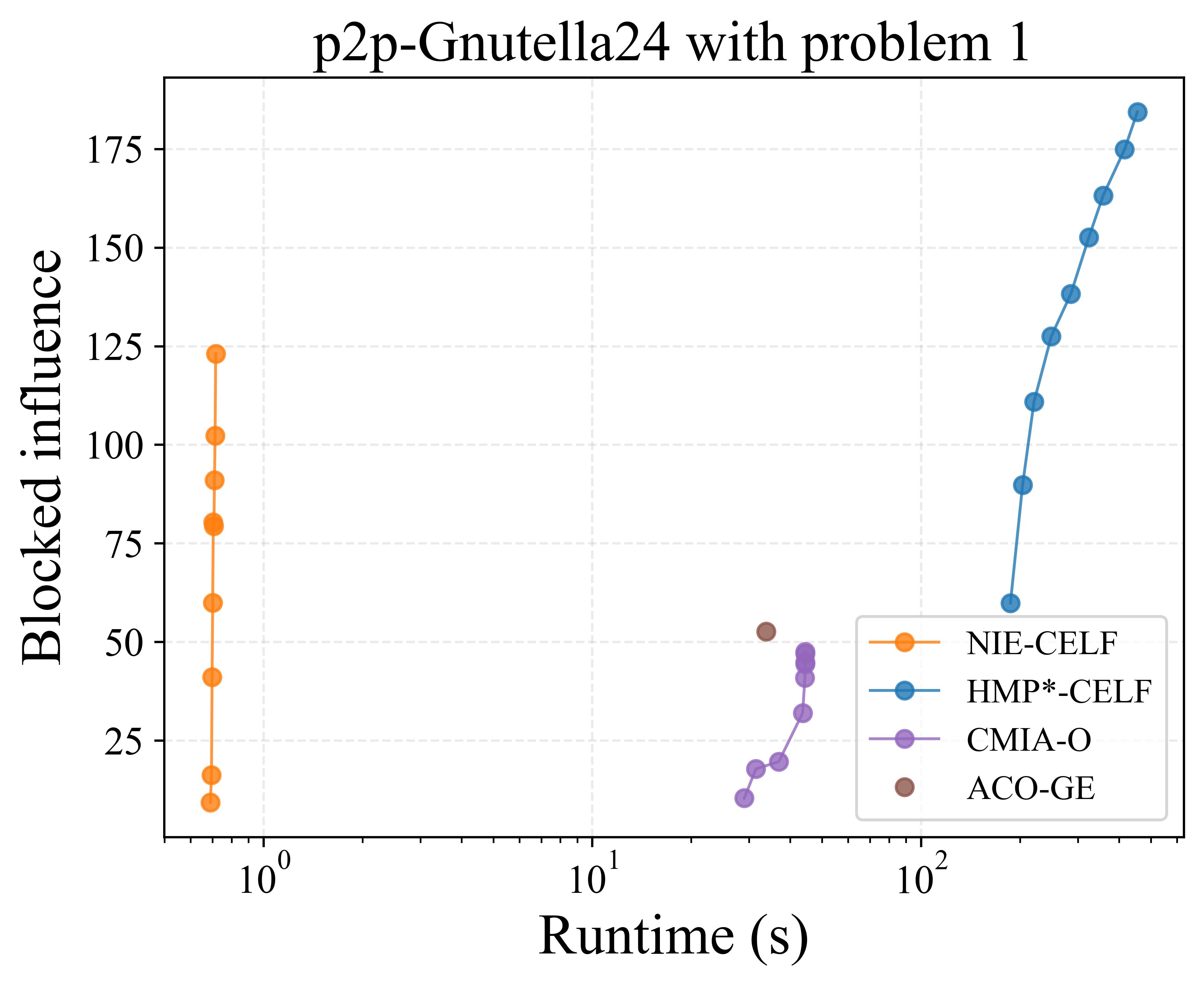}}
	\hfil
	\subfloat{\includegraphics[width=1.7in]{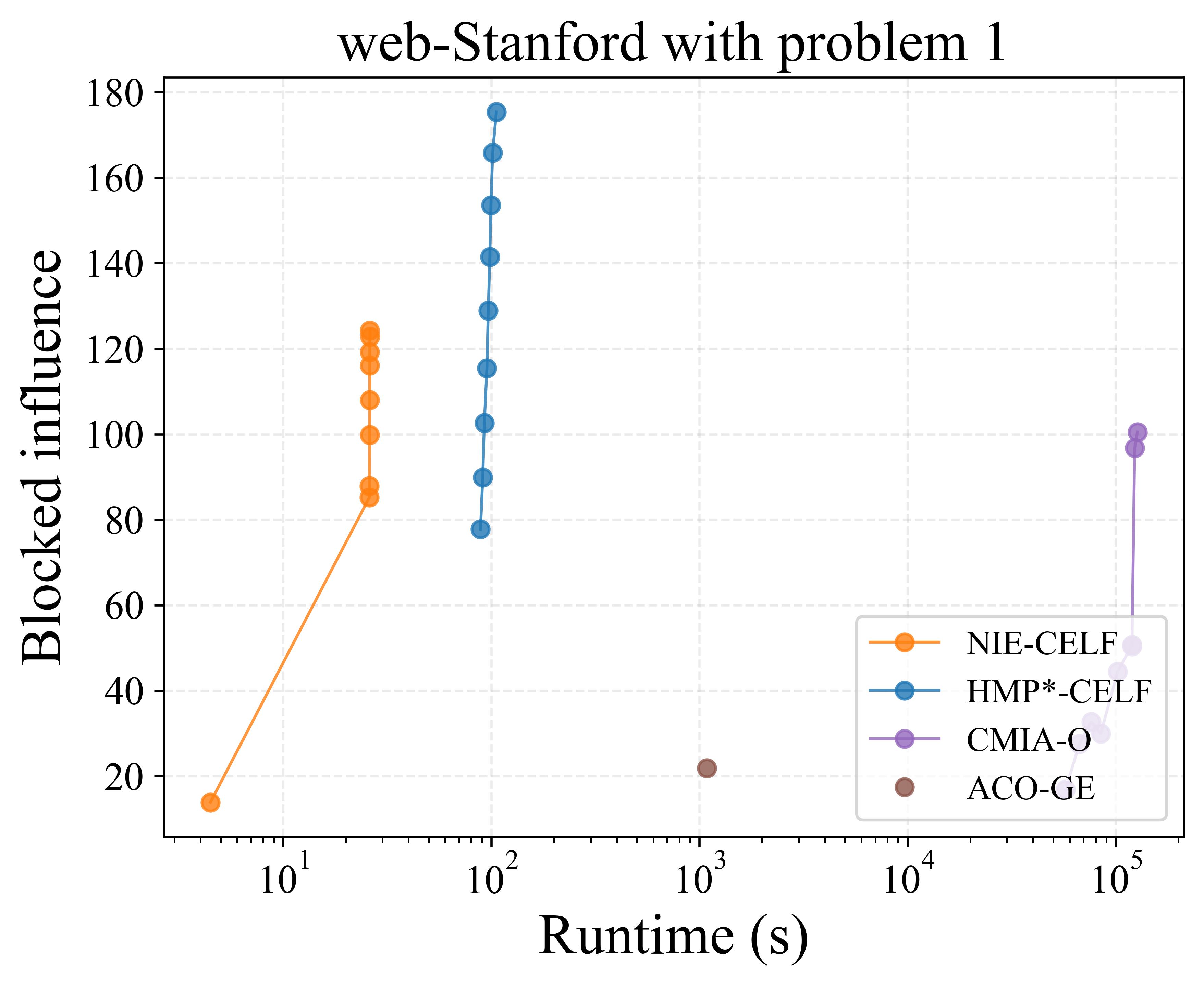}}
	\hfil\\\vspace{-2mm}
	\caption{Curves of the optimization quality (i.e. the blocked influence) vs. runtime of the optimization methods on five test problems, within a time budget of 48 hours. The result of ACO-GE is shown by one point because it is a rank-based algorithm rather than an iterative optimization algorithm.}
	\label{convergence curves}
\vspace{-4mm}
\end{figure*}

\subsection{Comparison in Terms of Runtime}
In this work, the most concerned performance indicator is the runtime $T$.
This section compares the runtimes of different methods given a target optimization quality.
The target quality is determined by the final objective value achieved by NIE-CELF.
Since it is challenging to ensure that two methods achieve the exactly same objective value during the optimization process, we adopt the following comparison approach that is slightly unfair to NIE-CELF.
That is, the compared method is considered to have reached the target quality if it, for the first time, discovers a solution whose quality is not worse than the target quality, and then the recorded runtime of its previous iteration is reported.
If the compared method fails to find such a solution, 
the runtime at which the method terminates is reported.

Table \ref{runtime} lists the runtime of the methods to achieve the target optimization quality.
The first observation is that NIE-CELF can solve the test problems mostly within one second.
For the web-Stanford network with more than 280,000 nodes and 2,300,000 edges, NIE-CELF can solve the test problems within 30 seconds.
These results demonstrate that NIE-CELF can achieve real-time solutions to IBM problems of practical scales.
Moreover, it is observed that all the compared methods consume more runtime than NIE-CELF on all the test problems, except on the second problem of the power-law graph network where NIE-CELF consumes slightly more runtime than CMIA-O.
In fact, some methods such as MCSs-CELF and StratLearner*-CELF even fail to reach the target optimization quality in 48 hours on the test problems with more than 10,000 nodes.
On each network, the average ratio between the runtime of the compared method and that of NIE-CELF, i.e., average speedup, is also presented in Table \ref{runtime}.
Notably, on all the five networks, NIE-CELF achieves substantial speedups over all the compared methods.
For example, compared with MCSs-CELF, NIE-CELF is at least four orders of magnitude faster on the test problems with more than 1,000 nodes.

Furthermore, we utilize the Mann-Whitney U test with a significance level $\alpha=0.05$ to determine whether there is a significant difference between the average runtime of NIE-CELF and that of the compared method on the same network.
The comparison results show that NIE-CELF is significantly better than all the comapred methods on all the networks, except on the smallest network power-law graph the difference between NIE-CELF and CMIA-O is not significant.
Nevertheless, NIE-CELF can scale up to large-scale networks much better than CMIA-O, since the runtime of the latter increases dramatically with the problem scale.
Overall, these observations clearly demonstrate the superiority of NIE-CELF to the compared methods, in terms of obtaining real-time solutions to large-scale IBM problems.

\subsection{Comparison in Terms of Optimization Quality}
In addition to runtime,  the other important performance metric is the optimization quality that can be achieved within a time budget.
Since this work focuses on real-time solutions, each method is given a time budget of one minute.

Table~\ref{quality} presents the optimization quality achieved by the methods.
The first observation is that NIE-CELF is the only method that can obtain solutions within one minute for all the test problems. 
Other methods fail to find real-time solutions for large-scale test problems. Similar to the previous experiments, we utilize the Mann-Whitney U test with a significance level $\alpha=0.05$ to determine whether there is a significant difference between the optimization quality of NIE-CELF and that of the compared method on the same network (note that the test is only conducted when the compared method has available results).
Compared with HMP*-CELF and CMIA-O, we find that although on some small-scale networks the compared methods can achieve better optimization quality, the quality difference between them and NIE-CELF is not significant. The comparison between NIE-CELF and ACO-GE shows that the quality difference is not significant for the p2p-Gnutella08 and p2p-Gnutella24 test problems. However, it is worth noting that NIE-CELF can achieve better solutions than ACO-GE, as the average quality ratio is larger than 1.

Furthermore, the performance of the methods is also tested using a much longer time budget of 48 hours.
Figure~\ref{convergence curves} plots the optimization quality of these methods along runtime (in log domain).
For the sake of brevity, for each network, only the results on one problem are illustrated.
Compared to other methods, NIE-CELF exhibits a significant advantage in terms of efficiency.
It can find high-quality solutions much faster than others, often by several orders of magnitude, though its final optimization quality might be lower than that of the other methods.
This strongly indicates the suitability of NIE-CELF for scenarios where a high-quality solution to IBM problems needs to be obtained promptly.

\begin{figure*}[tbp]
	\centering
	\subfloat{\includegraphics[width=1.7in]{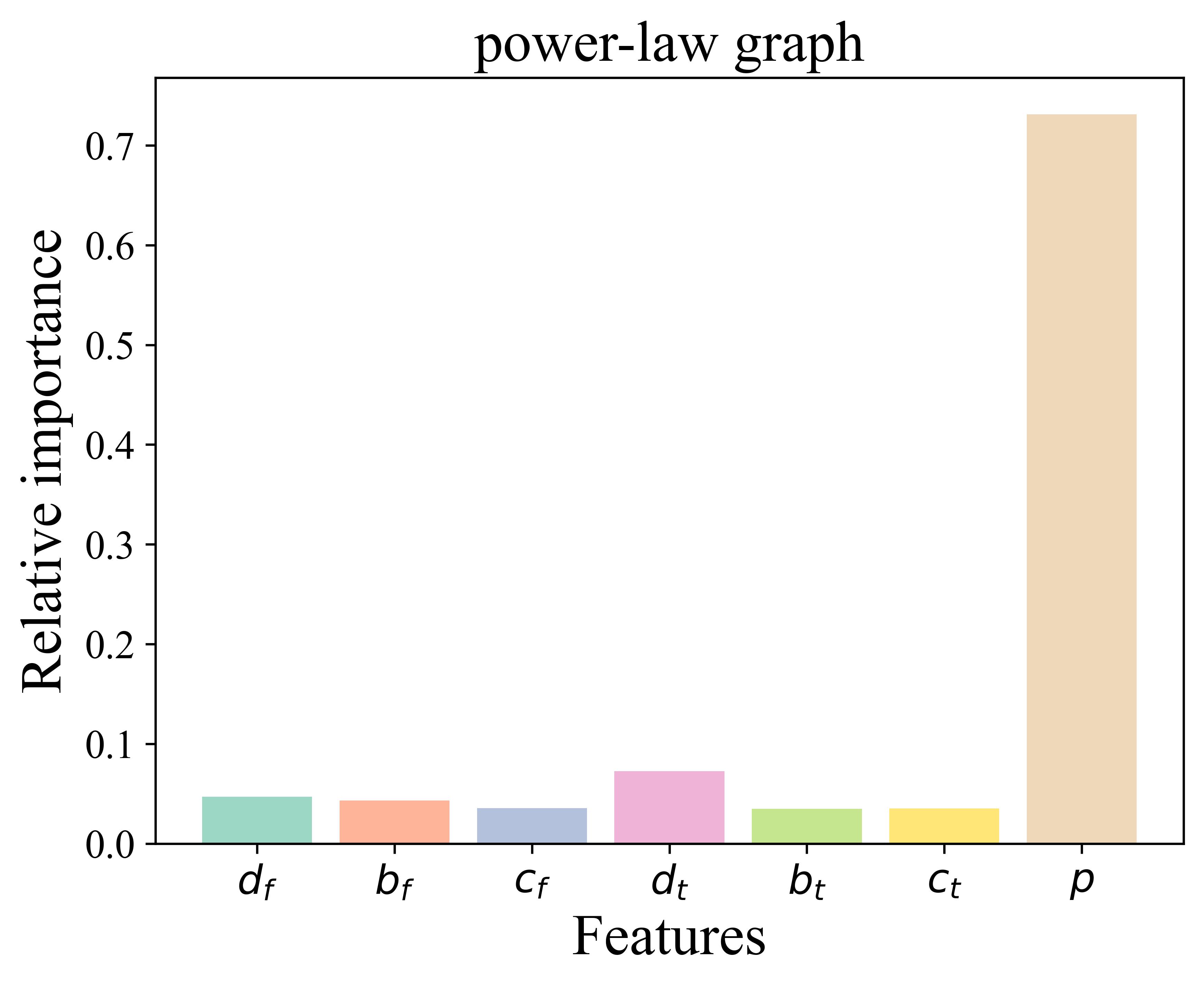}}
	\hfil
	\subfloat{\includegraphics[width=1.7in]{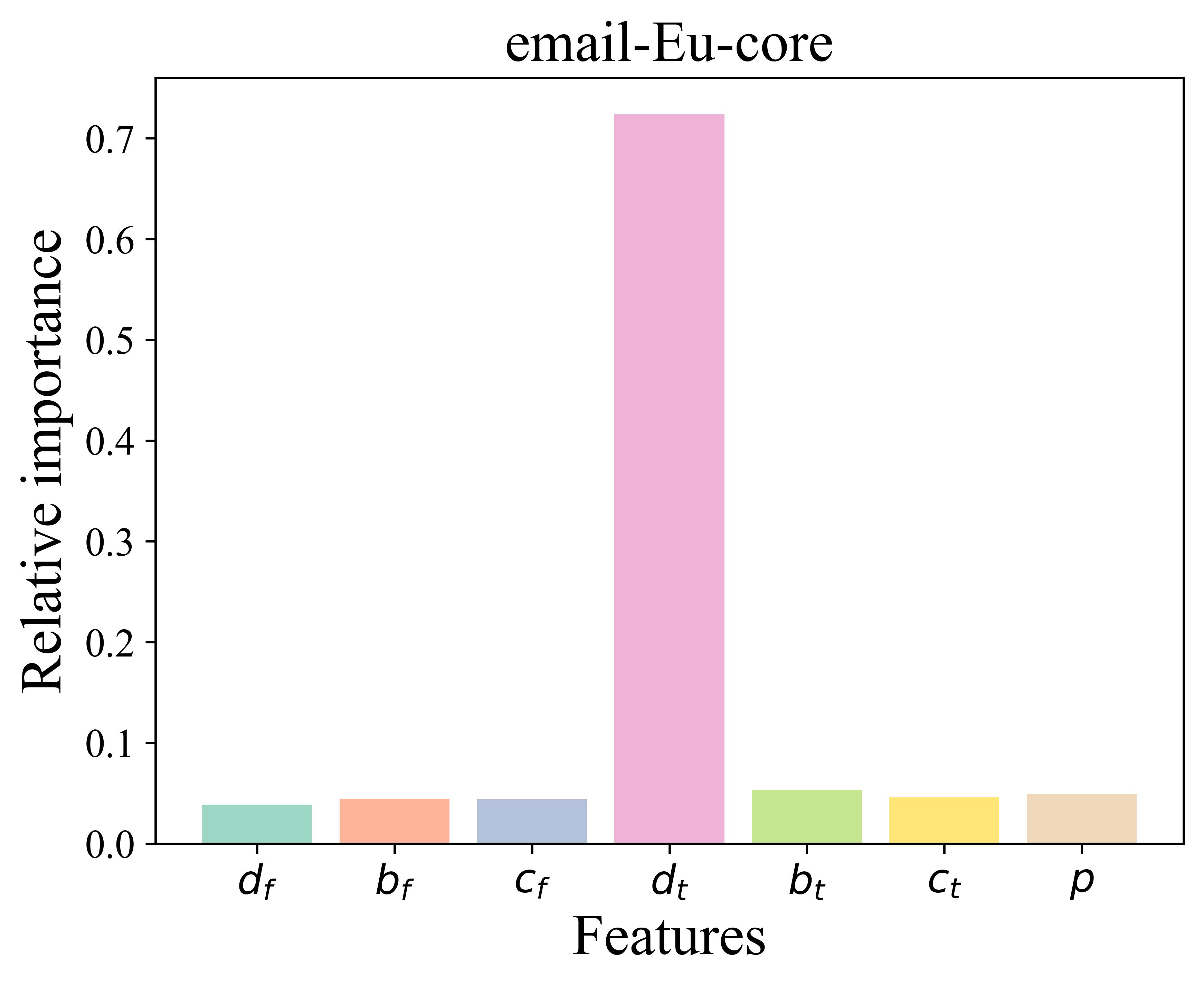}}
	\hfil
	\subfloat{\includegraphics[width=1.7in]{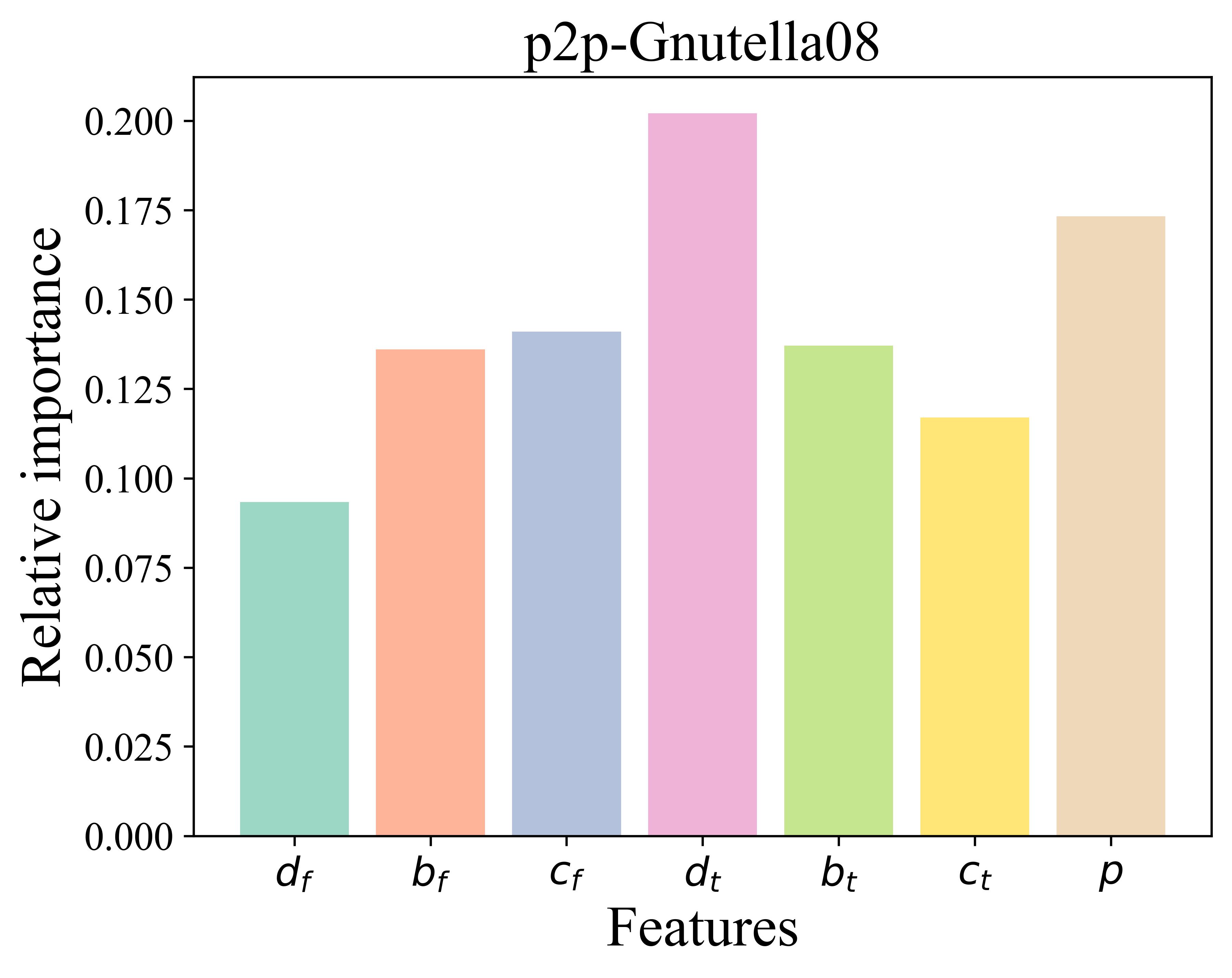}}
	\hfil \\ \vspace{-2mm}
	\subfloat{\includegraphics[width=1.7in]{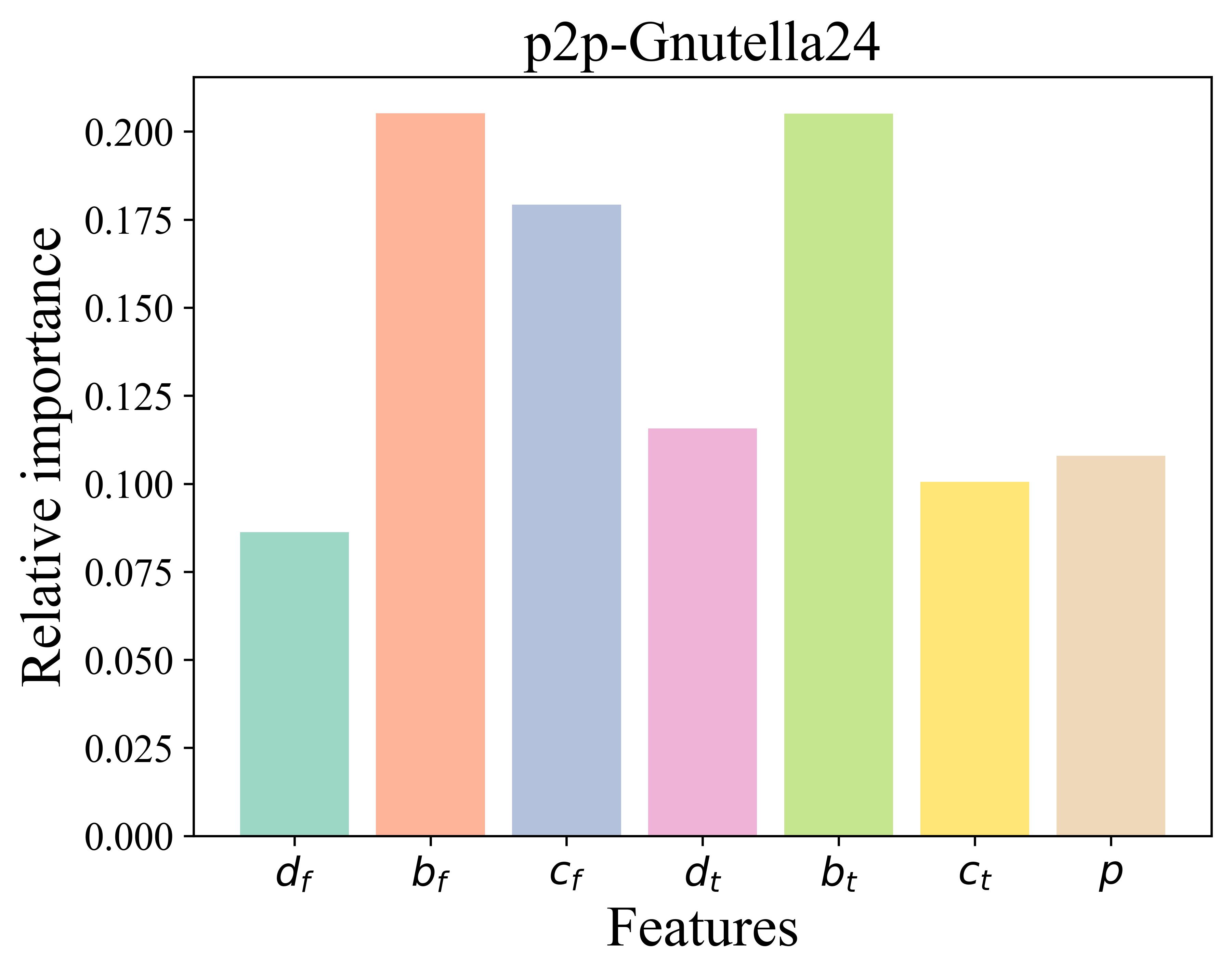}}
	\hfil
	\subfloat{\includegraphics[width=1.7in]{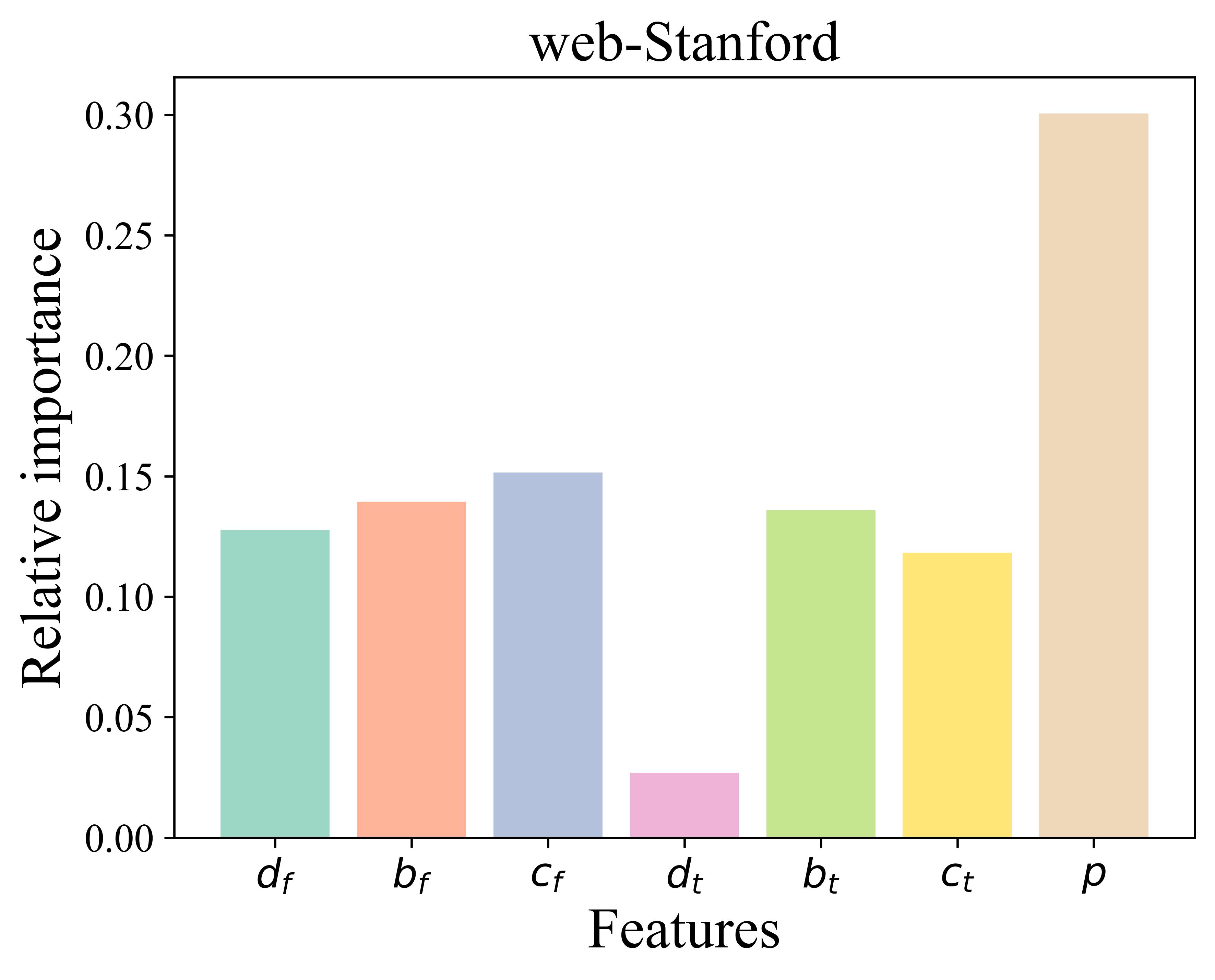}}
	\hfil\\\vspace{-2mm}
	\caption{{Feature importance analysis for NIE. $d_f$, $b_f$, and $c_f$ represents the neighborhood feature, the location feature, and the structure feature of $S_f$, respectively. $d_t$, $b_t$, and $c_t$ represents the neighborhood feature, the location feature, and the structure feature of $S_t$}, respectively. $p$ denotes the inter-relationship feature between $S_f$ and $S_t$.}
	\label{Feature Importance Analysis}
	\vspace{-4mm}
\end{figure*}

\begin{figure*}[tbp]
	\centering
	\subfloat{\includegraphics[width=1.7in]{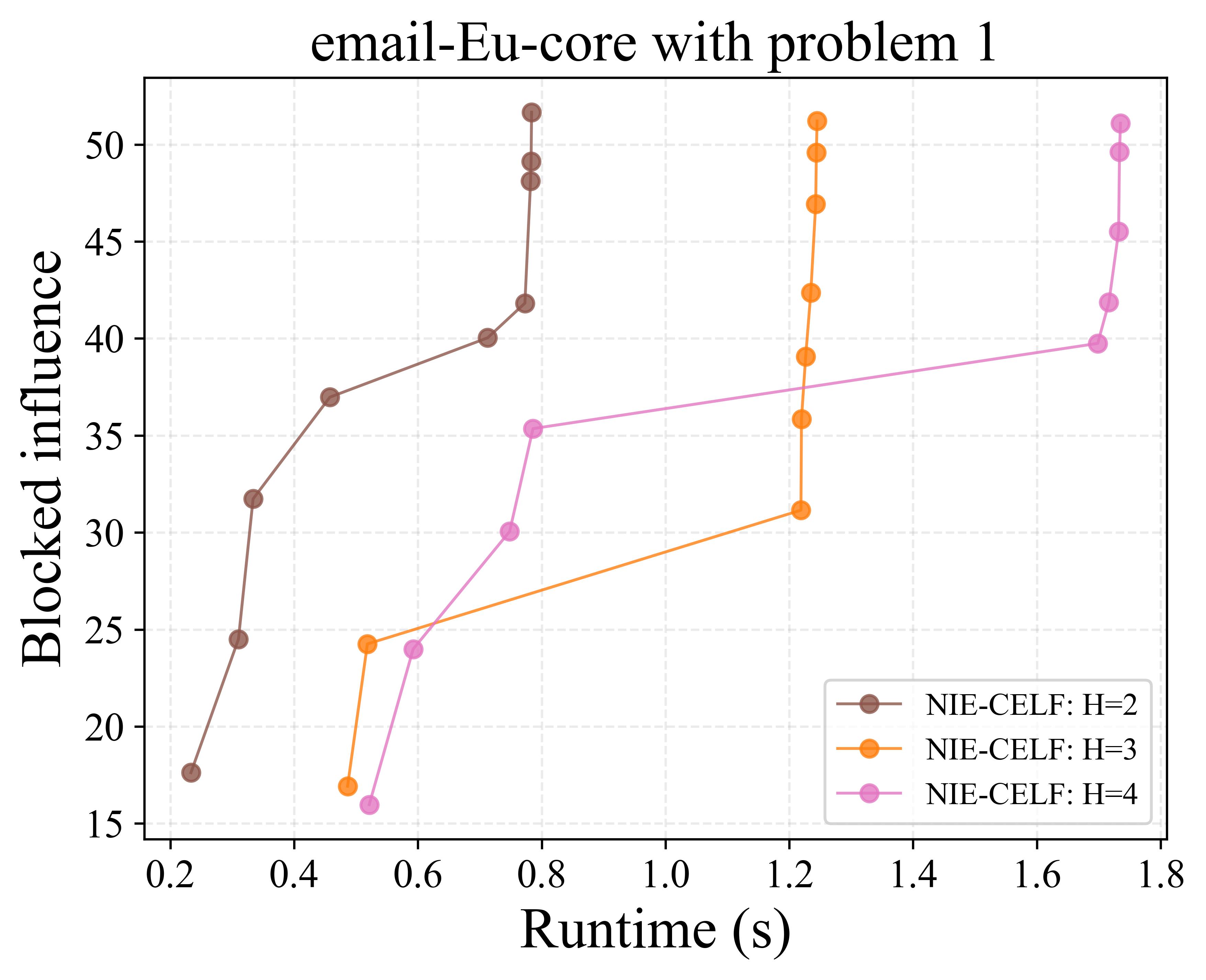}}
	\hfil
	\subfloat{\includegraphics[width=1.7in]{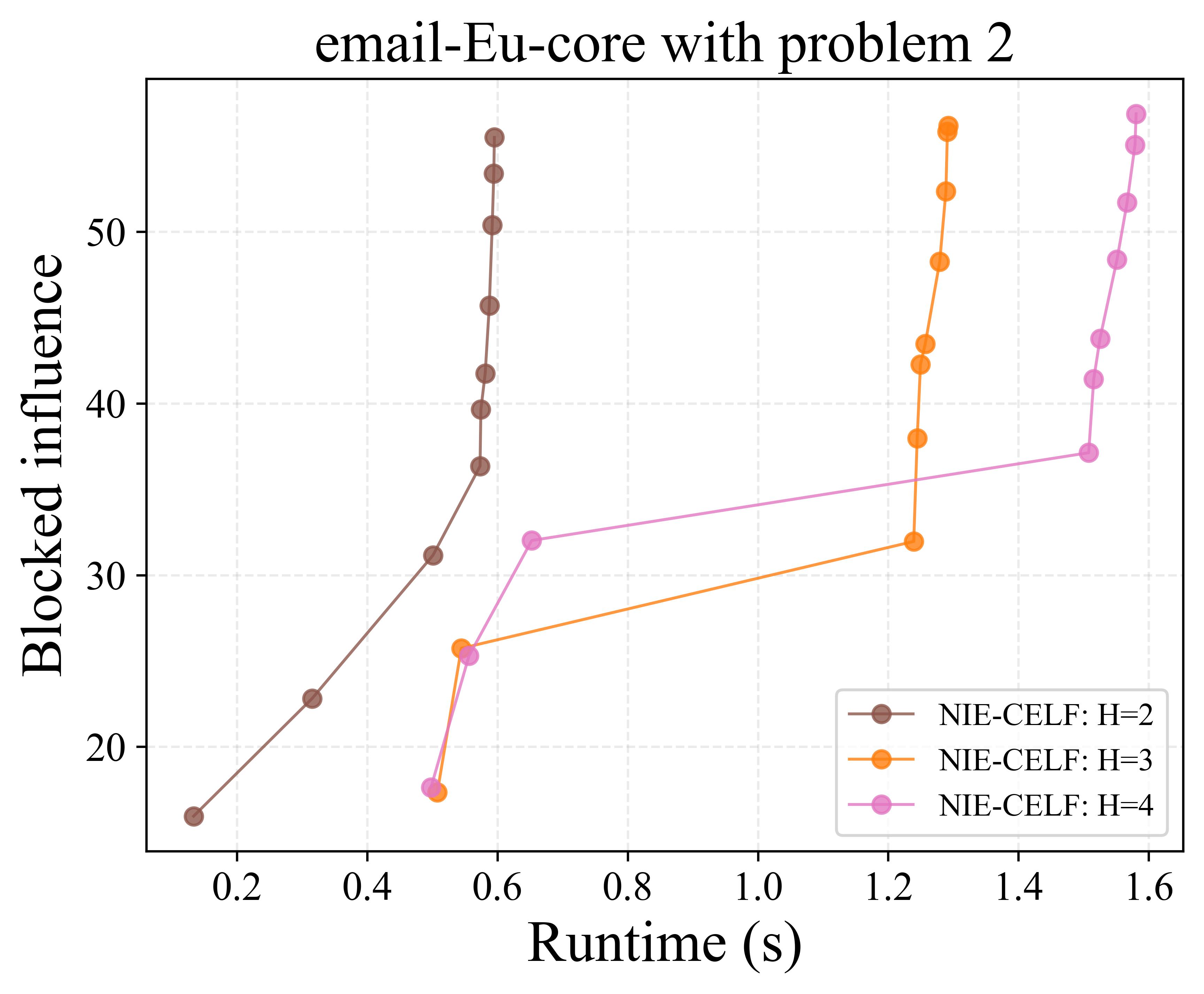}}
	\hfil
	\subfloat{\includegraphics[width=1.7in]{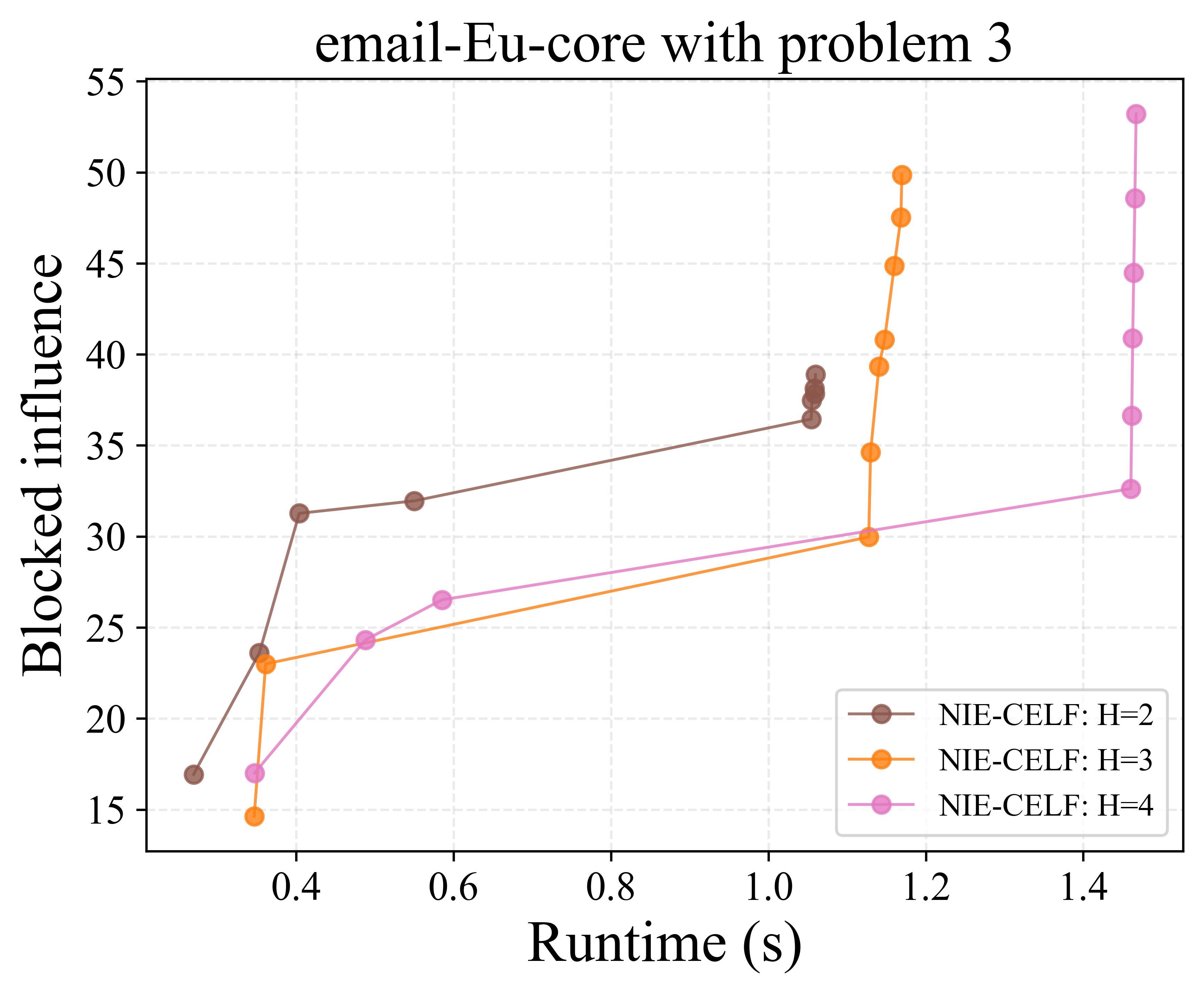}}
	\hfil
	\subfloat{\includegraphics[width=1.7in]{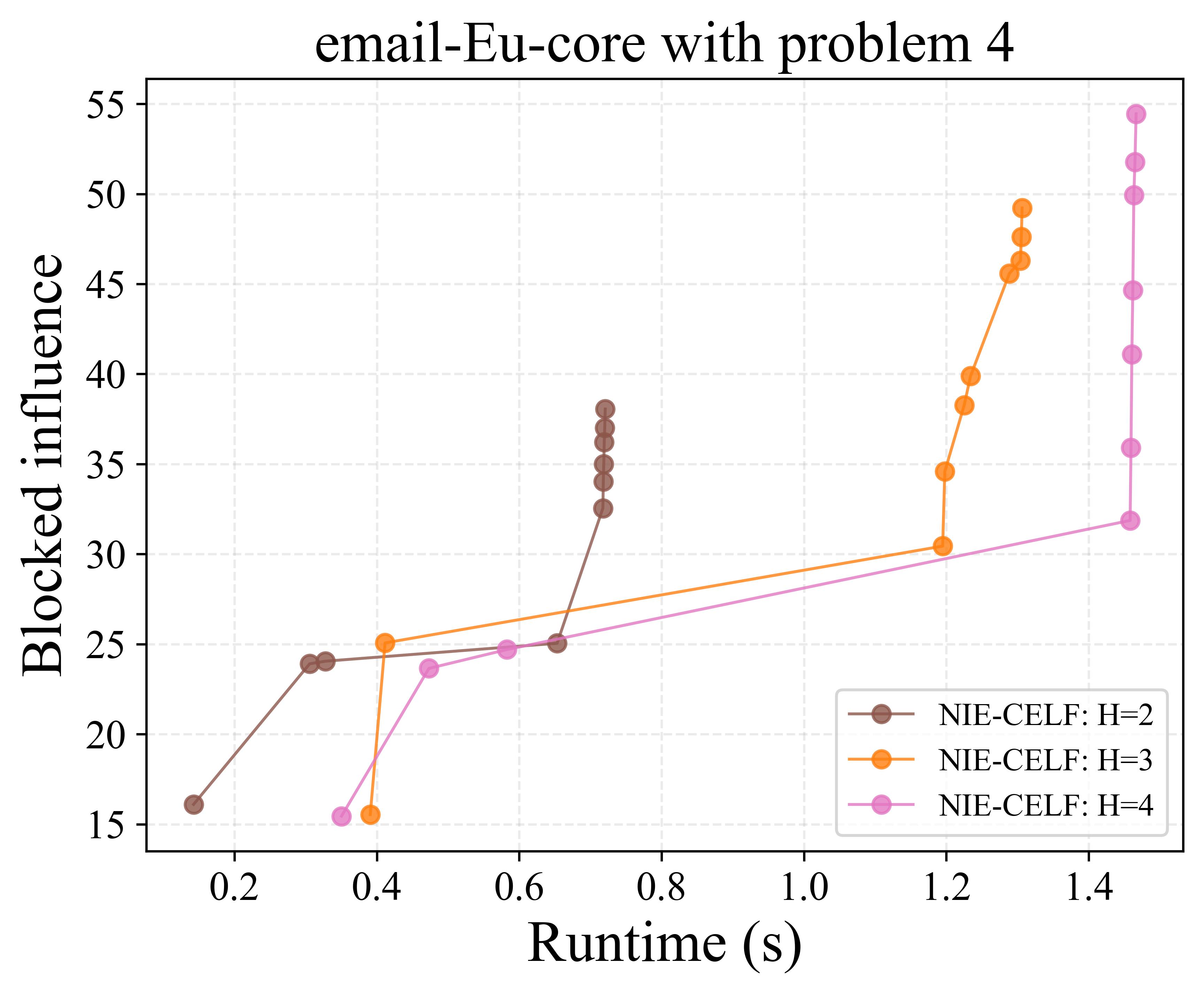}}
	\hfil\\ \vspace{-2mm}
	\subfloat{\includegraphics[width=1.7in]{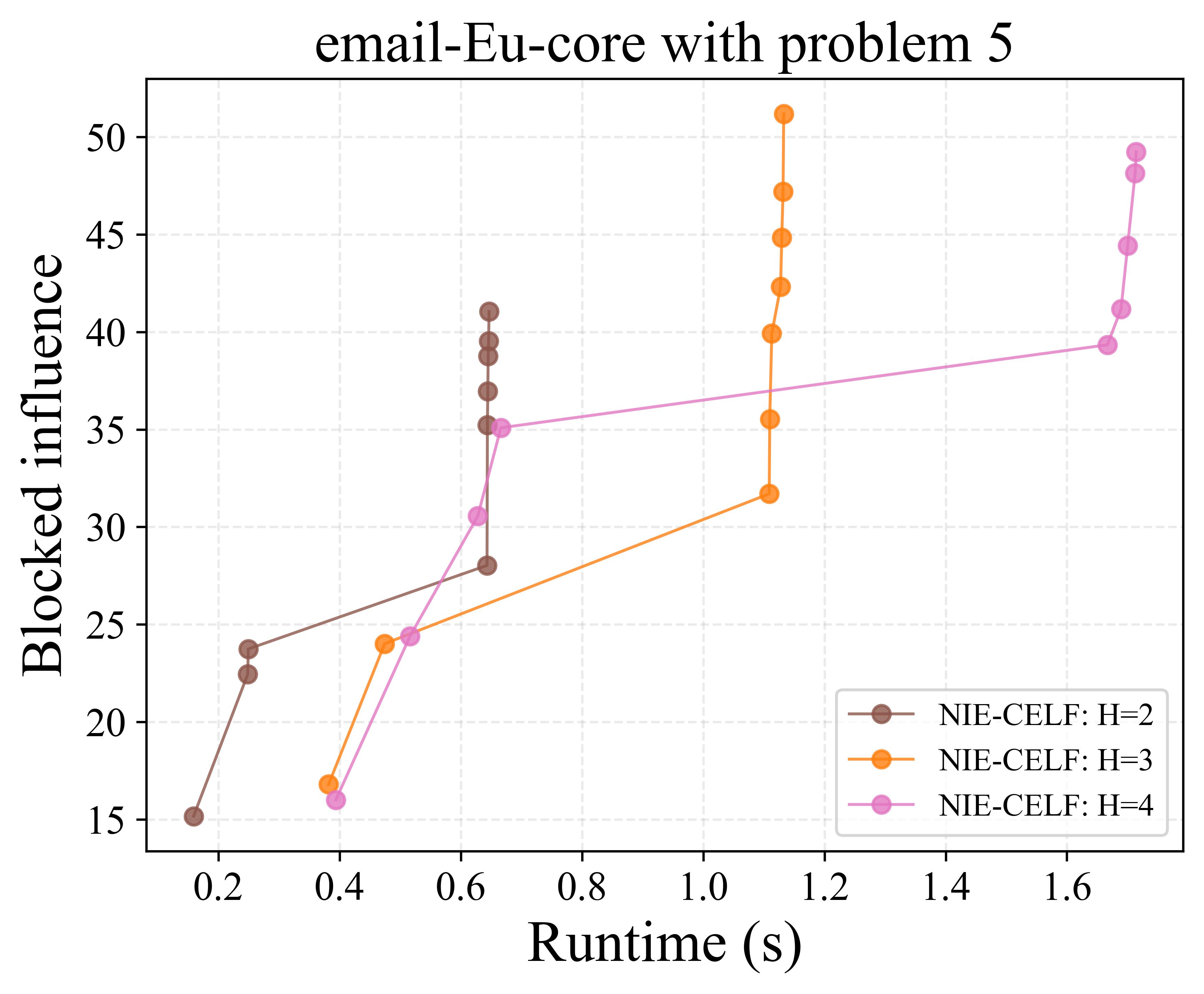}}
	\hfil
	\subfloat{\includegraphics[width=1.7in]{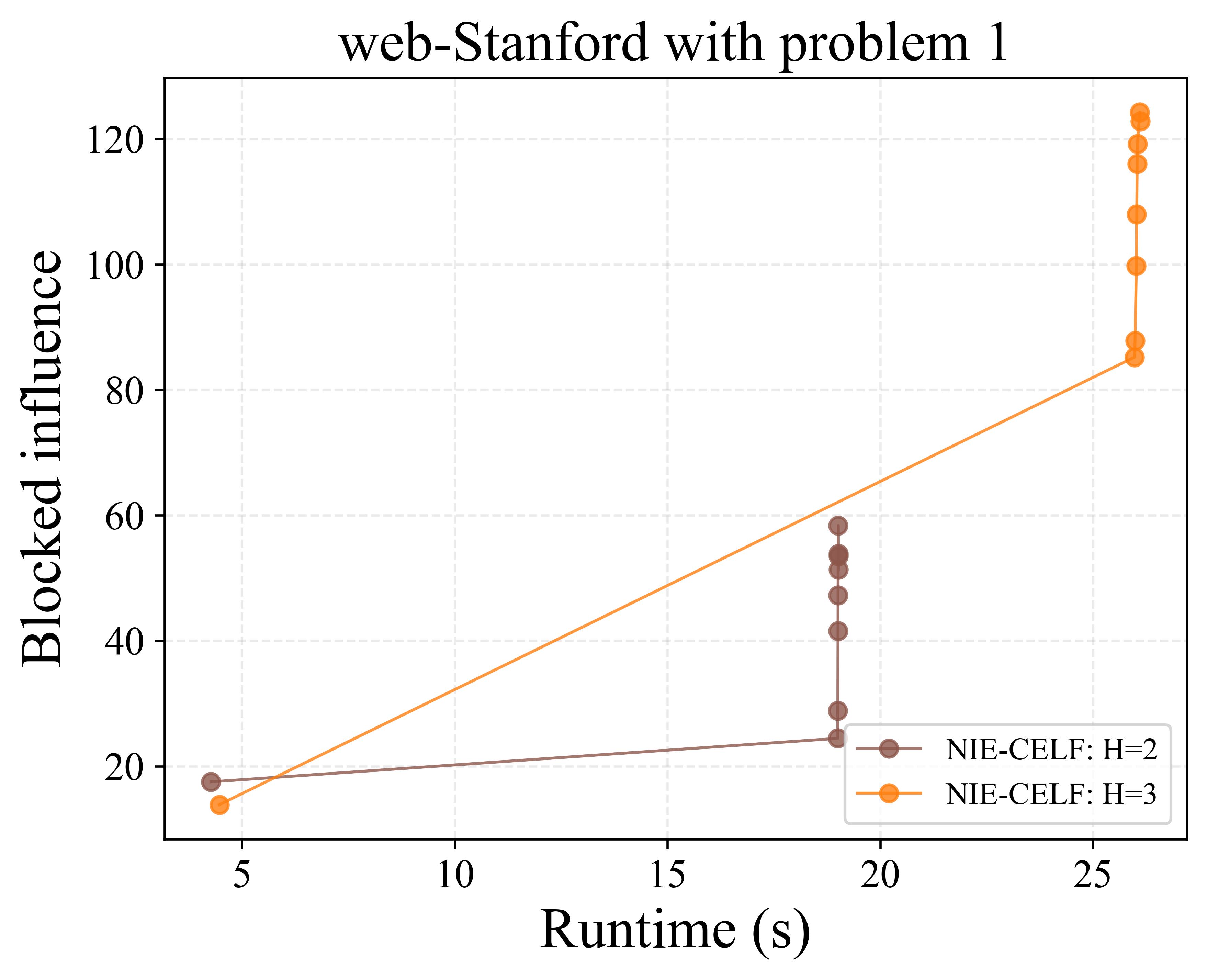}}
	\hfil
	\subfloat{\includegraphics[width=1.7in]{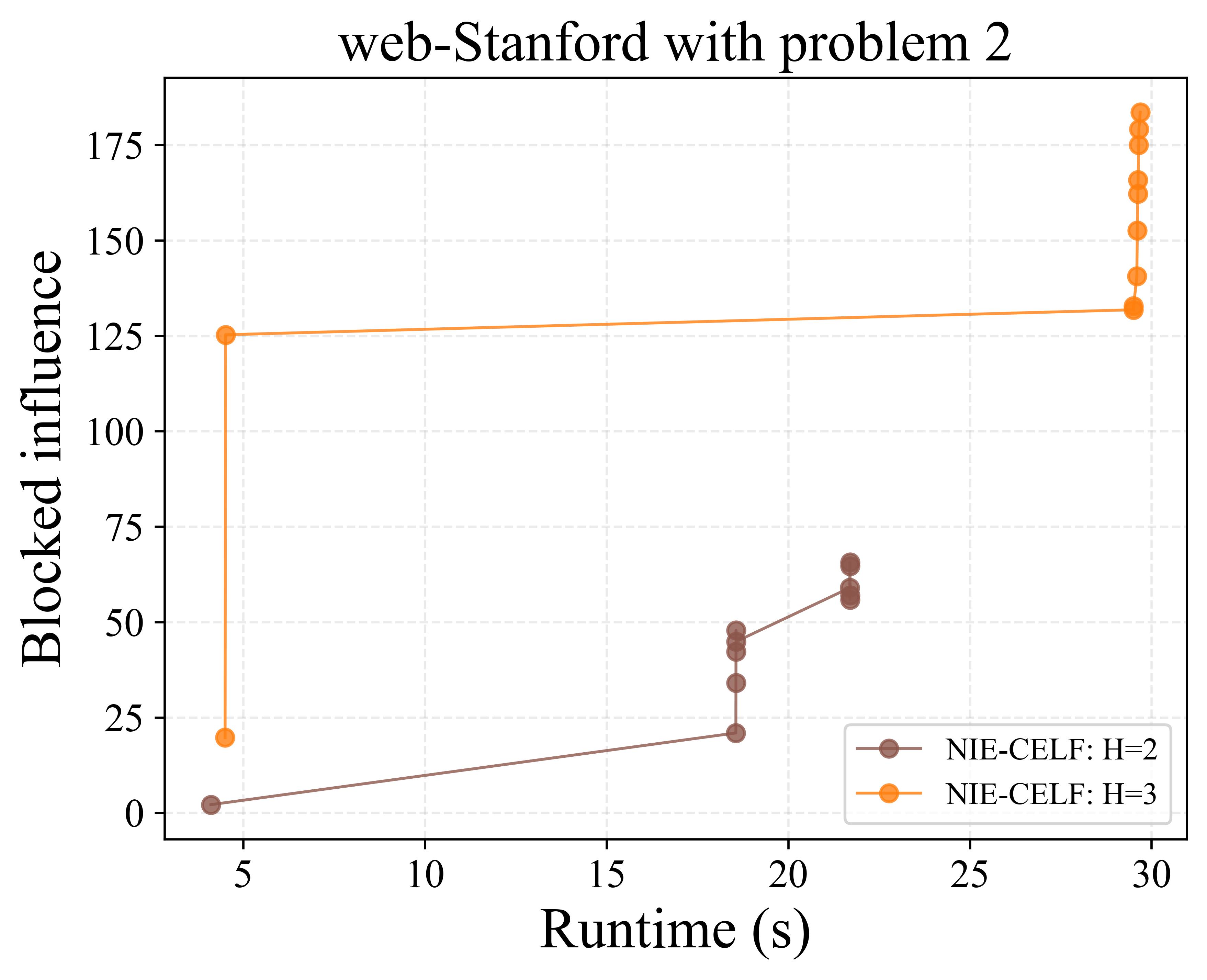}}
	\hfil
	\subfloat{\includegraphics[width=1.7in]{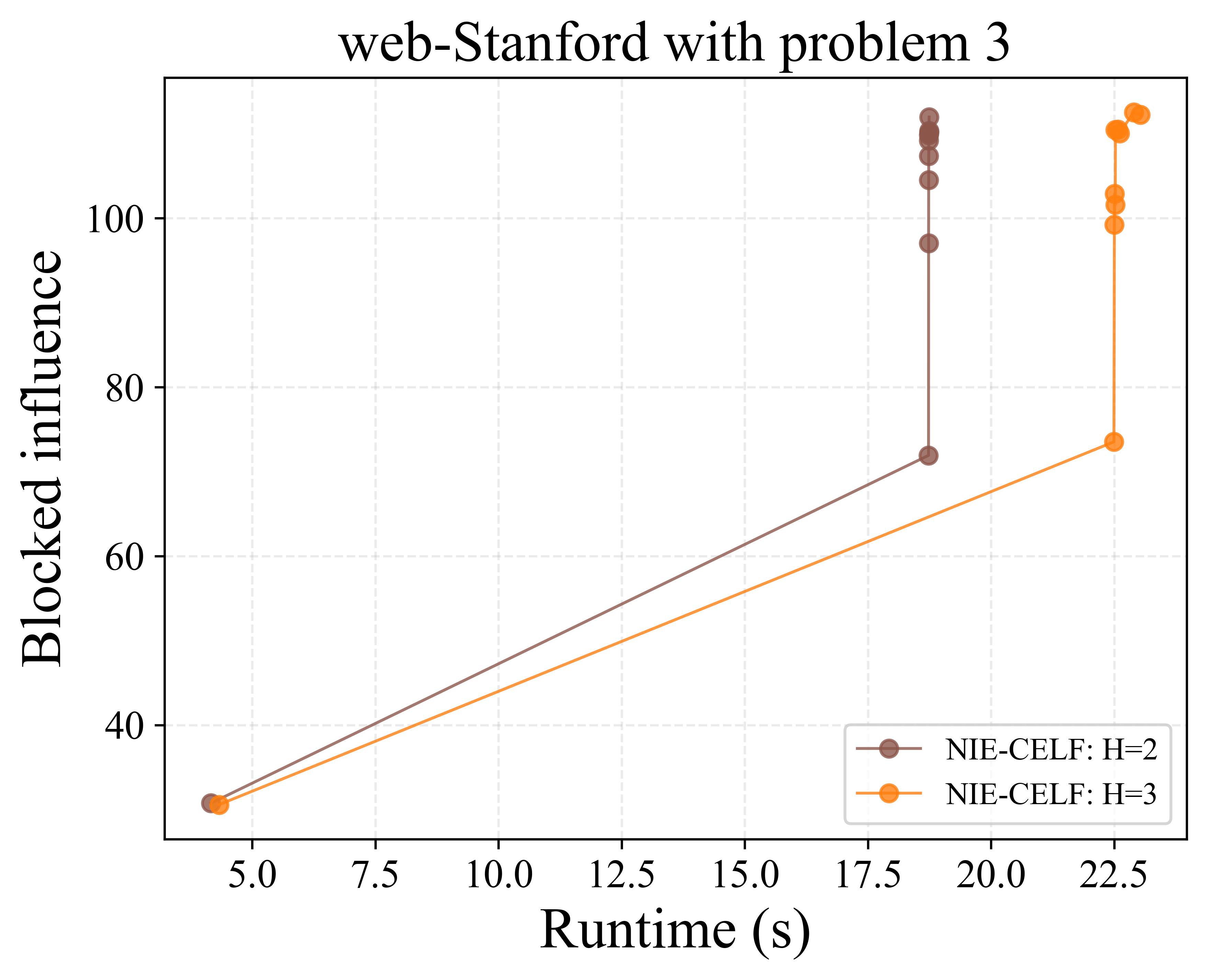}}
	\hfil\\ \vspace{-2mm}
	\subfloat{\includegraphics[width=1.7in]{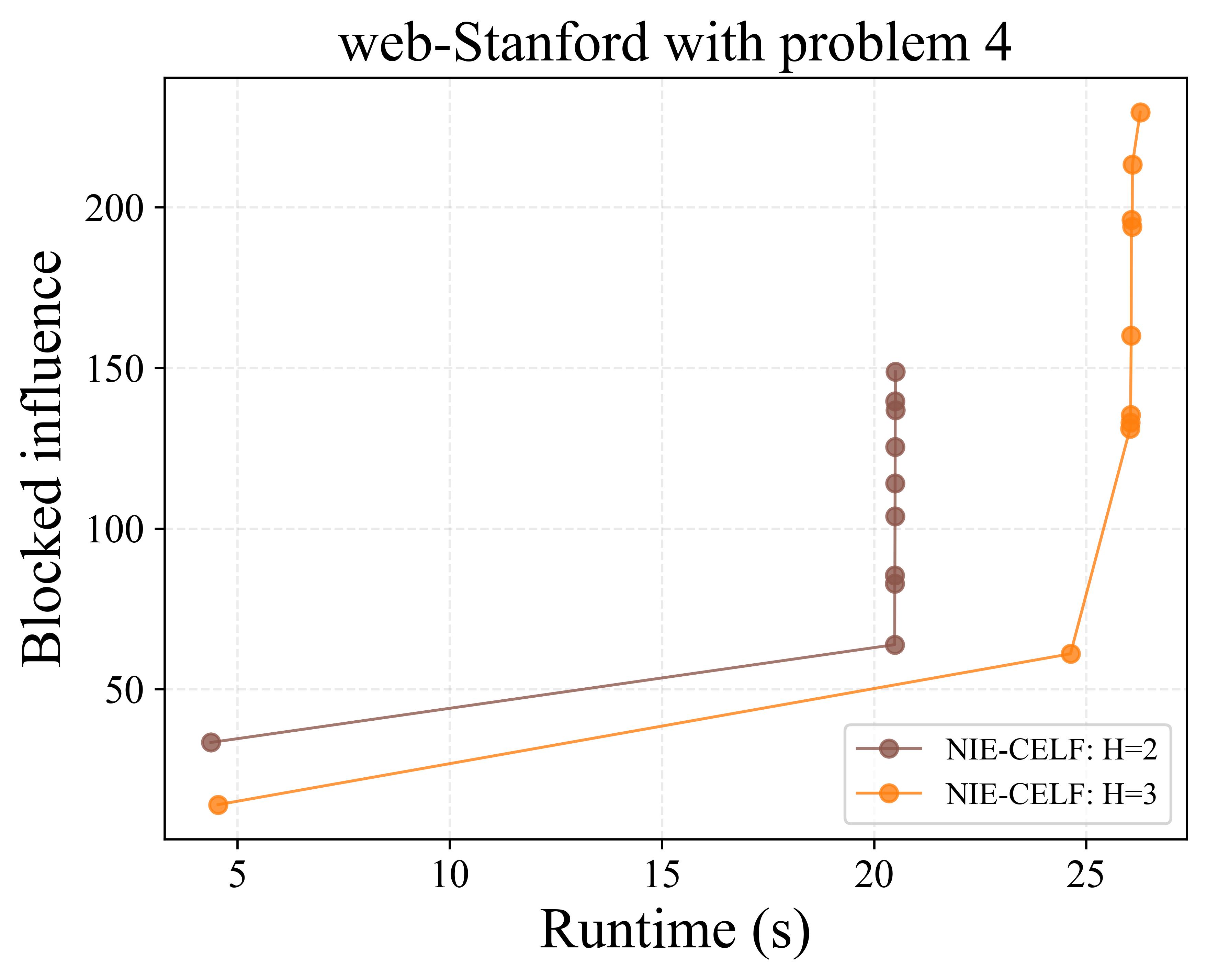}}
	\hfil
	\subfloat{\includegraphics[width=1.7in]{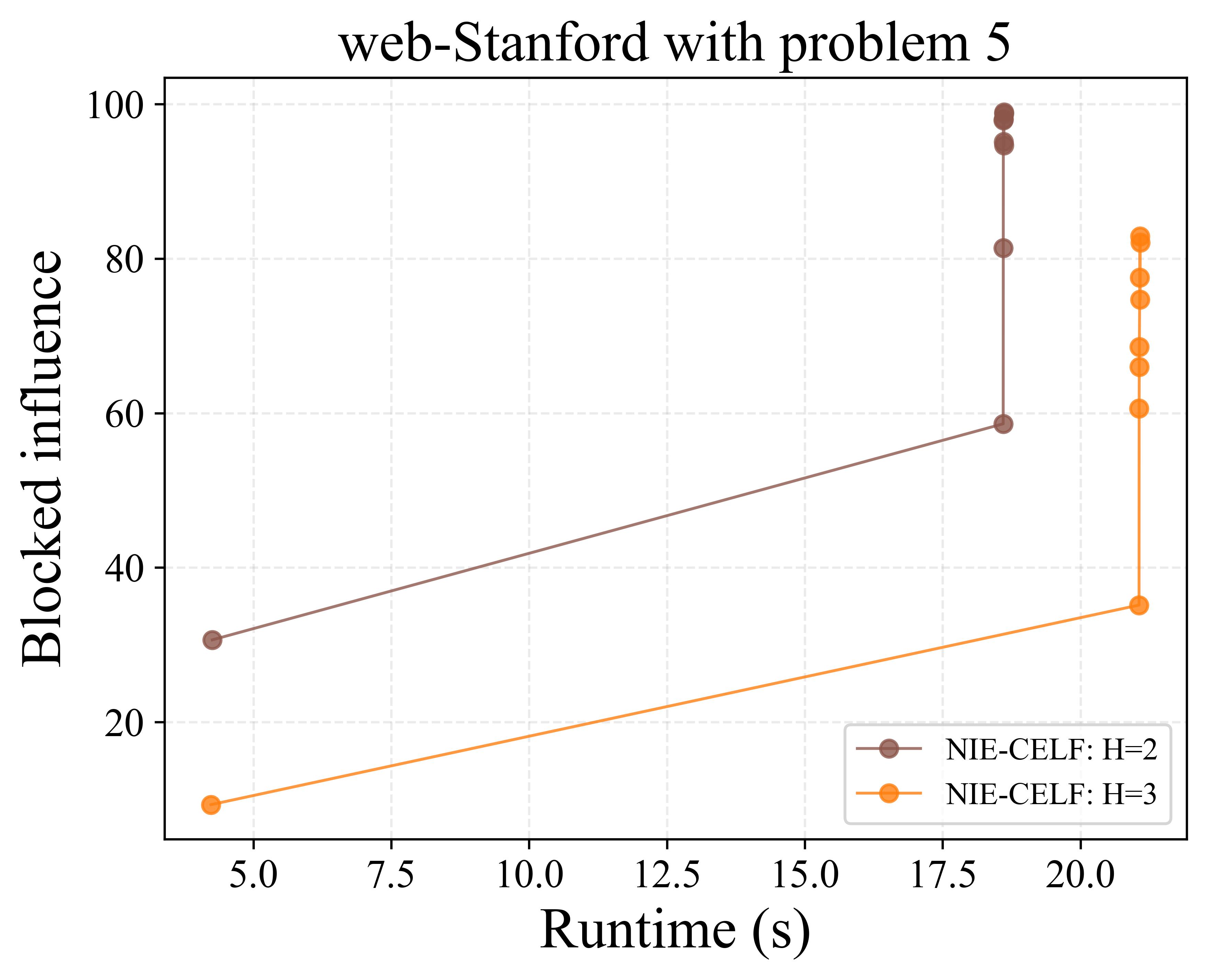}}
	\hfil \\ \vspace{-2mm}
	\caption{{The optimization quality (i.e. the blocked influence) achieved by NIE-CELF under different values of $H$.}}
	\label{Sensitivity Analysis 2}
 \vspace{-4mm}
\end{figure*}

\begin{figure}[tbp]
	\centering
	\subfloat{\includegraphics[width=1.8in]{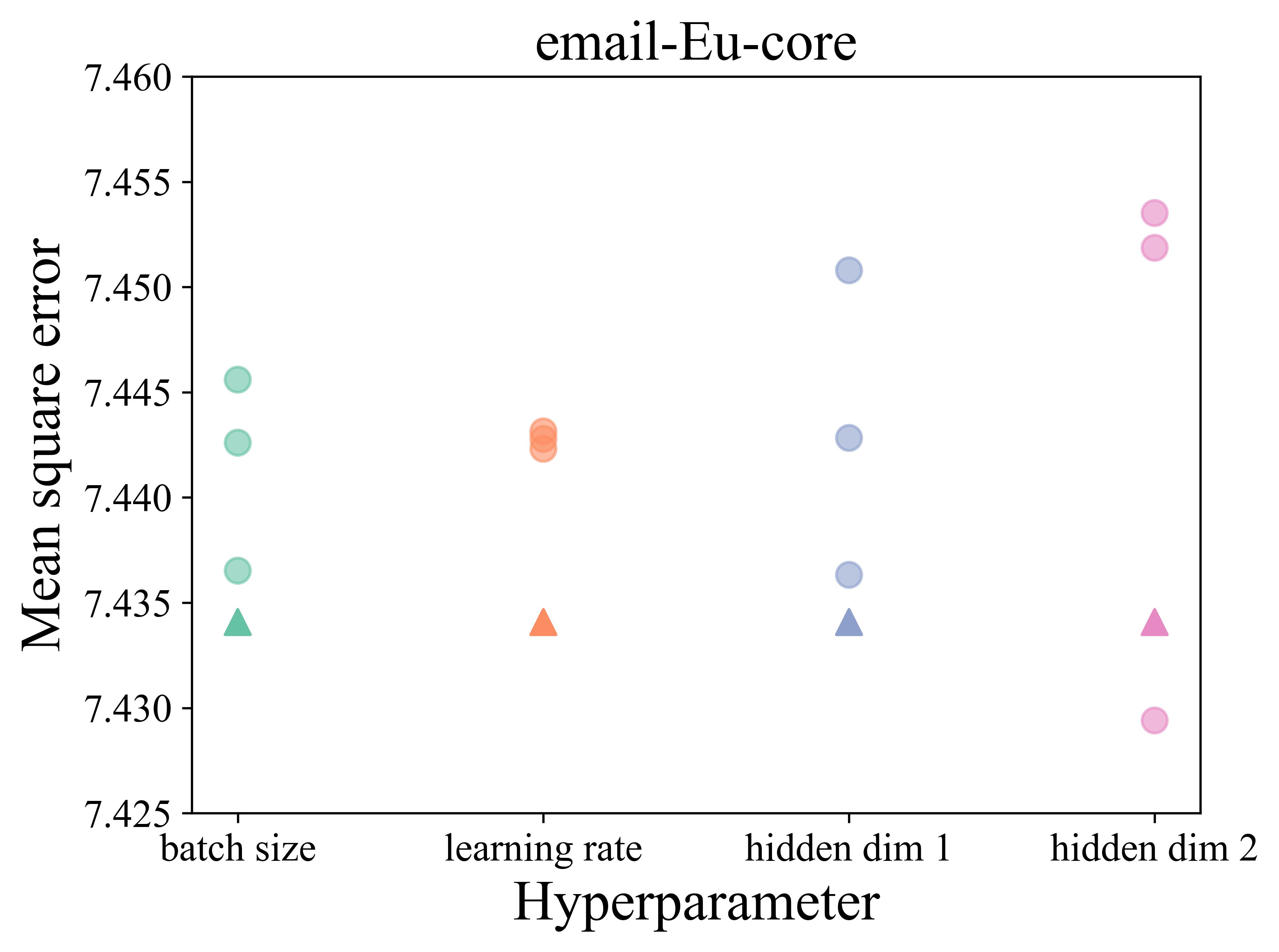}}
	\subfloat{\includegraphics[width=1.8in]{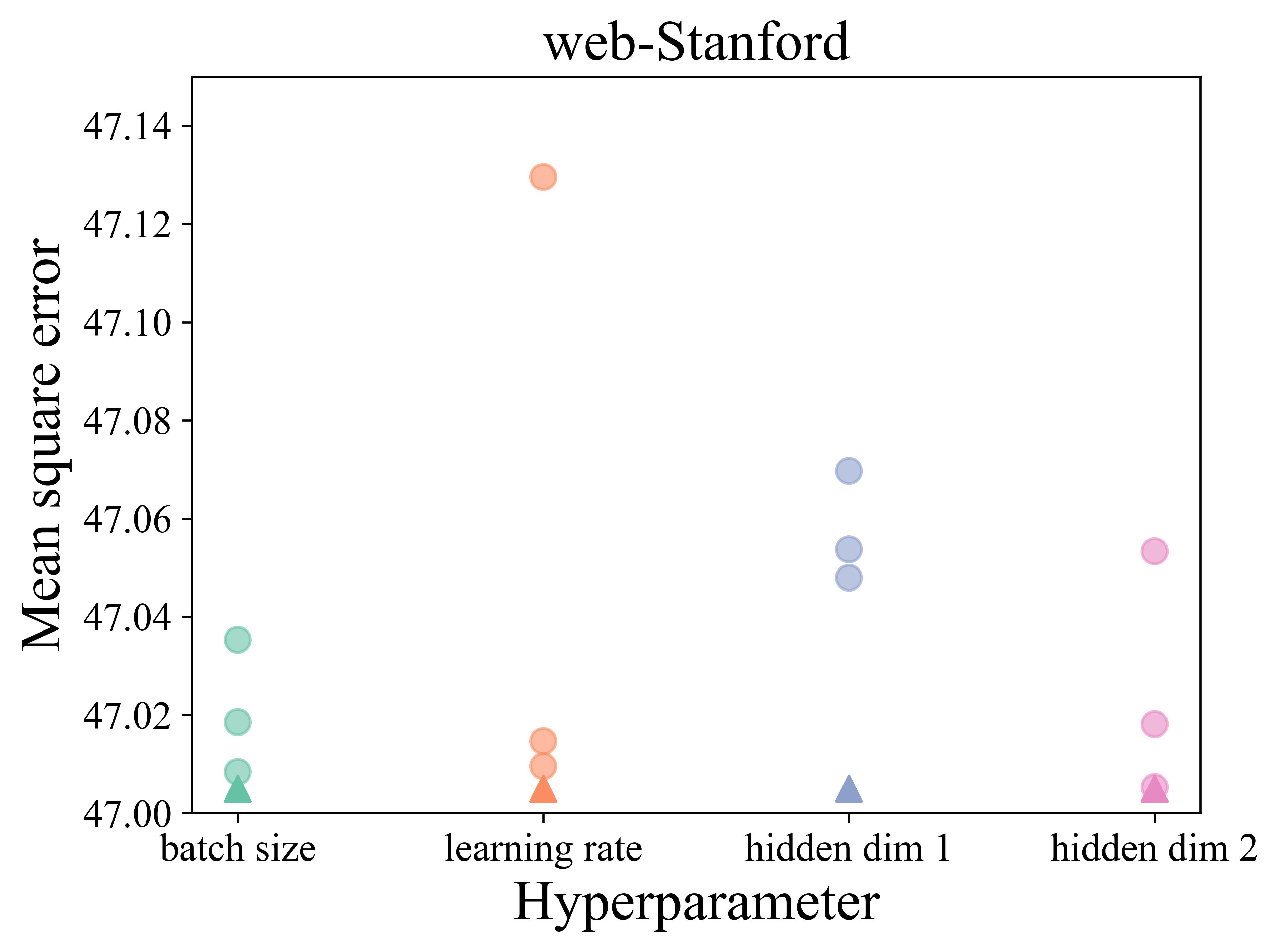}} 
	\hfil \\ \vspace{-2mm}
	\caption{{Performance of MLP under different hyperparameters values. The triangle represents the MSE of the baseline configuration, and circles represent the MSE of the modified configurations.}}
	\label{Sensitivity Analysis 1}
 \vspace{-4mm}
\end{figure}

\subsection{Feature Importance Analysis}
This section investigates the contributions of different features to NIE.
The relative importance score of each feature is obtained by calculating the normalized total reduction of impurity by feature (Gini importance) based on decision tree regression.
Figure~\ref{Feature Importance Analysis} illustrates the feature importance across all the networks.
It can be observed that the importance of features varies significantly across different networks.
For example, for p2p-Gnutella08 and p2p-Gnutella24, the seven features have similar importance, while for power-law graph and email-Eu-core, the inter-relationship feature $p$ and the location feature of $S_t$ (i.e., $d_t$) have significantly higher importance than other features, respectively.
Overall, all the features are important, and  it is recommended to employ all the features when building NIE for a new network.

\subsection{Choices of $H$ and Hyperparameters of MLP}
\label{sensitivity analysis}
This section discusses how to set $H$ and the hyperparameters of MLP. For the sake of brevity, we show the results on the email-Eu-core and web-Stanford which represent small-scale and large-scale networks, respectively. 

As shown in Section~\ref{complexity analysis}, larger $H$ can capture more graph information but requires more runtime. In practice, $H$ does not need to be too large because the well-known small world phenomenon reveals that two nodes are linked by short chains in the social network \cite{kleinberg2000small}. Therefore, we test $H\in\{2,3\}$ for web-Stanford and $H\in \{2,3,4\}$ for email-Eu-core. Figure~\ref{Sensitivity Analysis 2} confirms this observation and shows that larger $H$ leads to a better objective value but consumes more  runtime.
To balance optimization runtime and quality, we choose $H=3$ for web-Stanford and $H=2$ for other networks.

To set the hyperparamters of MLP, we start from a baseline hyperparameter configuration: batch size 512, learning rate 0.05, and a dimension of 128 for each hidden layer.
Then we vary one hyperparameter at a time and record the mean squared error (MSE) on an independent validation set. 
Specifically, a wide range of hyperparameter values are tested: batch size $\in\{128,256,512,1048\}$, learning rate $\in\{0.001,0.005,0.01,0.05\}$, dimension of the first hidden layer $\in \{64,128,256,512\}$, dimension of the second hidden layer $\in\{64,128,256,512\}$.
Figure~\ref{Sensitivity Analysis 1} displays the MSE under different hyperparameter configurations.
We find that the relative changes of MSE between the baseline configuration and the modified configurations are all less than 0.3\%.
Thus, MLP hyperparameters have minimal impact on its performance, allowing for consistent hyperparameter values across different networks.

\section{Conclusion}
\label{Conclusion}

This work focuses on achieving real-time solutions to IBM problems, which is crucial in practice.
A surrogate model, named NIE, is constructed as a fast approximation of the time-consuming MCSs by extracting multi-level information on the graph.
NIE can be readily combined with existing IBM optimization algorithms to improve efficiency. 
The complexity analysis and experimental results demonstrate that the NIE-based optimization method significantly outperforms the state-of-the-art methods in terms of both computational time and optimization quality (given a time budget of one minute).
Particularly, the NIE-based optimization method is suitable to cases where high-quality solutions to large-scale IBM problems need to be obtained in seconds. Our method can be generalized to graph-based combinatorial optimization problems where solutions need costly MCSs for evaluation, and their quality correlates with the topological patterns, such as the influence maximization problems and their variants.


There are some potential directions for future research.
Firstly, to achieve even higher efficiency, one possible method is to combine NIE with end-to-end solution prediction methods~\cite{LiuZTY23} or a portfolio of different optimization methods~\cite{LiuTY22}.
Secondly, to improve the solution quality, graph embedding might be employed to replace the current feature extractor in NIE.
Thirdly, it is interesting to investigate on utilizing graph contrastive learning~\cite{WuFCL0T22} to reduce the amount of labeled data for training NIE.
Fourthly, in real-world applications the diffusion model may be unknown or variable. This presents significant research potential for extending our approach to such scenarios.

\bibliographystyle{IEEEtran} 
\bibliography{library}

\end{document}